\documentclass{article}

\usepackage{PRIMEarxiv}

\usepackage[utf8]{inputenc} % allow utf-8 input
\usepackage[T1]{fontenc}    % use 8-bit T1 fonts
\usepackage{hyperref}       % hyperlinks
\usepackage{url}            % simple URL typesetting
\usepackage{booktabs}       % professional-quality tables
\usepackage{amsfonts}       % blackboard math symbols
\usepackage{nicefrac}       % compact symbols for 1/2, etc.
\usepackage{microtype}      % microtypography
\usepackage{lipsum}
\usepackage{fancyhdr}       % header
\usepackage{graphicx}       % graphics
\graphicspath{{media/}}     % organize your images and other figures under media/ folder

\usepackage{xcolor}
\usepackage{natbib}
\usepackage{amsmath}
\usepackage{multirow}
\usepackage{booktabs}
\usepackage{bbold}
\usepackage{soul}
\usepackage{amssymb}
\usepackage{rotating} 
\usepackage{pgf}
\usepackage{graphicx}
\usepackage{array} 

\newcommand*{\MaxNumber}{0.5}
\newcommand*{\Threshold}{0.05}
\newcommand{\cgr}[1]{%
  \pgfmathparse{ifthenelse(#1 < \Threshold, 1, 0)}
  \ifnum\pgfmathresult=1
    \pgfmathsetmacro{\PercentColor}{100*((\Threshold-#1))/\Threshold}
    \colorbox{green!\PercentColor}{$#1$}
  \else
    \pgfmathsetmacro{\PercentColor}{100*(#1-\Threshold)/(\MaxNumber-\Threshold)}
    \colorbox{red!\PercentColor}{$#1$}
  \fi
}
\newcommand{\BibTeX}{B\kern-.05em{\sc i\kern-.025em b}\kern-.08em\TeX}

\title{Evaluating LLMs Robustness in Less Resourced Languages with Proxy Models
}

\author{ \href{https://orcid.org/0009-0004-9251-972X}{\includegraphics[scale=0.06]{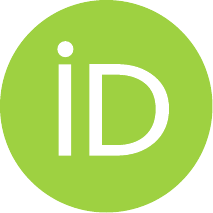}\hspace{1mm}Maciej~Chrabąszcz}\thanks{Equal contribution.}  \\
	NASK - National Research Institute,\\
	Warsaw, Poland \\
	\texttt{maciej.chrabaszcz@nask.pl} \\
	\And
	\href{https://orcid.org/0000-0002-5531-3499}{\includegraphics[scale=0.06]{orcid.pdf}\hspace{1mm}Katarzyna Lorenc }\footnotemark[1]\\
	NASK - National Research Institute,\\
	Warsaw, Poland \\
	\texttt{katarzyna.lorenc@nask.pl} \\
    \And
	\href{https://orcid.org/0000-0003-0617-7301}{\includegraphics[scale=0.06]{orcid.pdf}\hspace{1mm}Karolina Seweryn}\footnotemark[1] \\
	NASK - National Research Institute,\\
	Warsaw, Poland \\
	\texttt{karolina.seweryn@nask.pl} \\
}

\begin{document}
\maketitle

\begin{abstract}
Large language models (LLMs) have demonstrated impressive capabilities across various natural language processing (NLP) tasks in recent years. However, their susceptibility to jailbreaks and perturbations necessitates additional evaluations. Many LLMs are multilingual, but safety-related training data contains mainly high-resource languages like English. This can leave them vulnerable to perturbations in low-resource languages such as Polish. 
We show how surprisingly strong attacks can be cheaply created by altering just a few characters and using a small proxy model for word importance calculation. We find that these character and word-level attacks drastically alter the predictions of different LLMs, suggesting a potential vulnerability that can be used to circumvent their internal safety mechanisms.
We validate our attack construction methodology on Polish, a low-resource language, and find potential vulnerabilities of LLMs in this language. Additionally, we show how it can be extended to other languages. We release the created datasets and code for further research.
\end{abstract}

% keywords can be removed
% \keywords{First keyword \and Second keyword \and More}

\section{Introduction}
\begin{figure}[!t]
    \centering
    \includegraphics[width=0.5\linewidth]{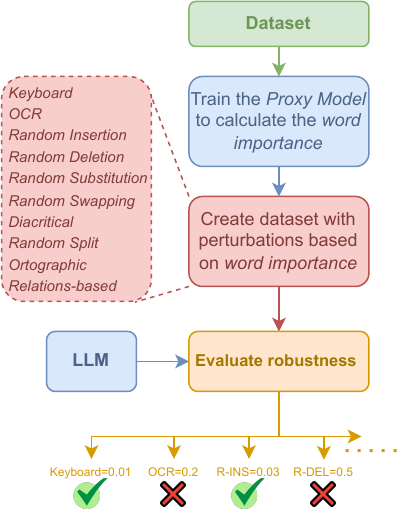}
    \caption{Overview of the proposed framework. We can calculate word importance using a trained proxy model on the desired datasets. Next, with word importance, we perturb the most important words with perturbations of interest. Then, we can evaluate the robustness of LLMs to those perturbations. By setting a threshold, we can create automatic robustness checks highlighting problems with the model during development.}
    \label{fig:framework_overview}
\end{figure}
Language models (LMs) \citep{devlin-etal-2019-bert, dubey2024llama3herdmodels} excel in natural language understanding (NLU) and natural language generation (NLG) tasks, enabling numerous everyday applications. However, recent studies \citep{Jin2019IsBR,Li2020BERTATTACKAA,wang-etal-2020-t3,zang-etal-2020-word, Wei2023Jailbroken} reveal their susceptibility to attacks that mimic human-like input perturbations. Therefore, it is crucial to test these models' robustness to perturbations.

LLMs research predominantly targets high-resource languages, e.g., English \citep{dubey2024llama3herdmodels,team2024gemma}. However, the emergence of multilingual language models \citep{Conneau2019UnsupervisedCR,commandr,yang2024qwen2technicalreport} introduces new vulnerabilities, particularly when incorporating lower-resourced languages. These languages pose a challenge due to limited data, making it difficult to enhance their multilingual robustness against even simple perturbations during fine-tuning processes, such as supervised fine-tuning (SFT) and model alignment. Therefore, assessing the robustness of multilingual models is crucial, especially for low-resource languages.

To address the issue of evaluating multilingual language models’ robustness, we propose a framework for generating perturbed datasets by utilizing proxy models and attribution methods. Those datasets can be used to assess LLMs' safety in terms of robustness, allowing developers to examine models' robustness to selected perturbations and lowering the risk of misleading their models with simple perturbations after release. Our framework leverages proxy models, which, combined with attribution methods, allow for the cheap identification of the most important words for a specific task, enabling the creation of evaluation examples by perturbing only the most important words. We validate our methodology on Polish and find potential vulnerabilities in LLMs in this language. This methodology can be easily adapted to other languages with minimal linguistic effort. By performing targeted perturbations on important words, we can rigorously test and ensure the robustness of various language models to perturbations of interest.

\begin{table*}[ht!]
\centering
\small 
\caption{Examples illustrating different perturbations that successfully changed models' predictions.}
\resizebox{\textwidth}{!}{ % To scale the table to the page width
\begin{tabular}{l|l|p{6.5cm}|c}
\toprule
\textbf{Modification} & \textbf{Dataset} & \textbf{Sample (\st{Strikethrough} = Original Text, \textcolor{red}{red} = Perturbation)} & \textbf{Label $\rightarrow$ Prediction} \\ \midrule
\multirow{2}{*}{OCR} & \multirow{2}{*}{AR} & \textbf{PL}: \st{słaba} \textcolor{red}{sła6a} jakość, typowa podróbka, nie polecam & \multirow{2}{*}{1 $\rightarrow$ 2} \\ %\cline{4-4}
 & & \multicolumn{1}{p{6.5cm}|}{\textbf{EN}: Poor quality, typical fake, I do not recommend.} & \\ \hline
\multirow{2}{*}{Rel} & \multirow{2}{*}{AC} & \textbf{PL}: klient akceptuje \st{niniejszy} \textcolor{red}{następujący} regulamin & \multirow{2}{*}{0 $\rightarrow$ 1} \\ %\cline{4-4}
 & &  \multicolumn{1}{p{6.5cm}|}{\textbf{EN}: client accepts these terms and conditions.} & \\ \hline
\multirow{2}{*}{R-Sw} & \multirow{2}{*}{AC} & \textbf{PL}: Oświadczam, że otrzymałem, \st{zapoznałem} \textcolor{red}{zapzonałme}\ się i akceptuję & \multirow{2}{*}{1 $\rightarrow$ 0} \\ %\cline{4-4}
 & & \multicolumn{1}{p{6.5cm}|}{\textbf{EN}: I declare that I have received, read, and accept.} & \\ \hline
\multirow{2}{*}{Split} & \multirow{2}{*}{P-O} & \textbf{PL}: dzięki \st{niemu} \textcolor{red}{ni emu} jeszcze jestem studentem ; p & \multirow{2}{*}{plus m $\rightarrow$ minus s} \\ %\cline{4-4}
 & & \multicolumn{1}{p{6.5cm}|}{\textbf{EN}: thanks to him, I'm still a student ;p} & \\ \hline
\multirow{2}{*}{Key} & \multirow{2}{*}{P-O} & \textbf{PL}: UNIKAC nie \st{polecam} \textcolor{red}{poleFXm} . . . brak slow ( tz sa ale post mi zlikwiduja ) & \multirow{2}{*}{minus m $\rightarrow$ zero} \\ %\cline{4-4}
 & & \multicolumn{1}{p{6.5cm}|}{\textbf{EN}: AVOID, I don't recommend... I'm at a loss for words (well, I have words, but they'll delete my post).} & \\ \hline
\multirow{2}{*}{R-Del} & \multirow{2}{*}{AC} & \textbf{PL}: sąd \st{właściwy} \textcolor{red}{włściy} dla siedziby powoda. & \multirow{2}{*}{0 $\rightarrow$ 1} \\ %\cline{4-4}
 & & \multicolumn{1}{p{6.5cm}|}{\textbf{EN}: The court competent for the plaintiff} & \\ \hline
\multirow{2}{*}{R-Sub} & \multirow{2}{*}{CBD} & \textbf{PL}: @anonymized\_account Bierz tego @anonymized\_account razem \st{jesteście} \textcolor{red}{jesteśwZe} mocni & \multirow{2}{*}{0 $\rightarrow$ 1} \\ %\cline{4-4}
 & & \multicolumn{1}{p{6.5cm}|}{\textbf{EN}: @anonymized\_account take @anonymized\_account together, you are strong.} & \\ \hline
\multirow{2}{*}{R-Ins} & \multirow{2}{*}{P-O} & \textbf{PL}: Krotko : cala grupa zaliczyla , nie oddal nikomu ani jednego sprawka . Oceny \st{marzenie} \textcolor{red}{mqarzenJie} . Tyle : - ) & \multirow{2}{*}{plus m $\rightarrow$ zero} \\ %\cline{4-4}
 & & \multicolumn{1}{p{6.5cm}|}{\textbf{EN}: Shortly: the whole group passed, didn't hand in a single assignment to anyone. Dream grades, just a dream. That's it : - )} & \\ \hline
\multirow{2}{*}{Ort} & \multirow{2}{*}{AC} & \textbf{PL}: Za ewentualne zniszczenia odpowiada \st{opiekun} \textcolor{red}{opiekón} & \multirow{2}{*}{0 $\rightarrow$ 1} \\ %\cline{4-4}
 & & \multicolumn{1}{p{6.5cm}|}{\textbf{EN}: The supervisor is responsible for any potential damage.} & \\ \hline
\multirow{2}{*}{Dia} & \multirow{2}{*}{AC} & \textbf{PL}: Po opuszczeniu parkingu firma reklamacji nie \st{uwzględnia} \textcolor{red}{uwzglednia}. & \multirow{2}{*}{0 $\rightarrow$ 1} \\ %\cline{4-4}
 & & \multicolumn{1}{p{6.5cm}|}{\textbf{EN}: After leaving the parking lot, the company does not accept any complaints.} & \\ \midrule \bottomrule
\end{tabular}
}
\label{table:ape_examples}
\end{table*}

Our contributions are as follows: 
\begin{itemize} 
    \item We introduce a framework that generates human-understandable perturbed examples to assess the LLMs' robustness to perturbations.
    \item We curate Polish datasets, train proxy models on them, and conduct perturbations, resulting in the creation of the Polish dataset, which can be used to evaluate LLMs' robustness in Polish.
    \item We perform an extensive assessment of the robustness of LLMs, utilizing the created dataset. We identify the perturbations that result in the most substantial performance degradation. The insights gained from this analysis can assist model developers in improving the robustness of their models.
\end{itemize}

\section{Related Work}
% \clearpage

\subsection{Safety and Robustness}

Ensuring the robustness of AI models is crucial for maintaining model safety. Recent studies have demonstrated that LMs can be vulnerable to perturbations, leading to incorrect class predictions~\citep{Jin2019IsBR,li2019textbugger,zang-etal-2020-word} or the generation of undesirable text, even when the models are well-aligned~\citep{zou2023universal}. These findings underscore the necessity of evaluating models' robustness to perturbations. 

When generating a perturbed example, it is essential to preserve the original meaning of the text, which can be done by adding typos~\citep{li2019textbugger}, synonyms~\citep{Jin2019IsBR}, BERT-based substitutions~\citep{Li2020BERTATTACKAA}. Consequently, there has been a focus on developing adversarial datasets, such as AdvGLUE for English data~\citep{wang2021adversarial}, which has been extended for LLMs in the DecodingTrust framework~\citep{wang2023decodingtrust}. However, to the best of our knowledge, there are no comparable datasets available for Polish and many other lower-resourced languages.

\subsection{Attribution Methods}
Attribution methods aim to identify the most influential parts of the input for model predictions. While previous works on attacks~\citep{li2019textbugger,Li2020BERTATTACKAA} have calculated word importance by observing changes in predictions when a word is omitted or simple saliency attributions~\citep{simonyan2014deep,li2019textbugger}, these approaches can be problematic due to those methods being unfaithful to the model.

Alternative attribution methods were developed to offer more robust attribution in black and white-box scenarios. In black-box scenarios, where the model's internal mechanisms are inaccessible, methods such as SHAP (Shapley Additive explanations)~\citep{scott2017unified} and LIME (Local Interpretable Model-agnostic Explanations)~\citep{Ribeiro2016WhySI} help determine word importance.

In white-box scenarios, where model internals are accessible, gradient-based attribution methods have gained prominence. Notable examples include Grad x Input~\citep{Shrikumar2016NotJA}, Integrated Gradients~\citep{Sundararajan2017AxiomaticAF}, and SmoothGrad~\citep{Smilkov2017SmoothGradRN}. These approaches leverage the model's gradient information to quantify the contribution of each input feature to the final prediction.

The widespread adoption of transformer-based models in NLP has led to the development of attribution methods specifically designed for this architecture. These include Attention Rollout~\citep{Abnar2020QuantifyingAF} and other attention-based techniques~\citep{Chefer2021GenericAE,Chefer_2021_beyond_attention}.
\subsection{Language Modeling}
% napisac o modelach jezykowuych
Language modeling is a fundamental task in NLP that involves predicting the probability distribution of words or tokens in a sequence. These models have evolved significantly, starting from simple statistical approaches like n-grams to the more advanced neural network architectures that dominate the field today. Transformer-based language models like GPT~\citep{brown2020languagemodelsfewshotlearners} and BERT~\citep{devlin-etal-2019-bert} have achieved state-of-the-art results in various tasks, including text classification and generation. 

In Polish, several transformer-based models have been developed~\citep{mroczkowski-etal-2021-herbert, Kleczek2020polbert,dadas2024assessing}. Among these, Bielik~\citep{ociepa2024bielik} currently stands as the most prominent Polish-dedicated generative model, though some multilingual models also provide support for the Polish language (LLama3.1~\citep{dubey2024llama3herdmodels}, OpenChat~\citep{wang2023openchat}, CommandR~\citep{commandr}). % cohere, bielik, llama, 
%The core objective of these models is to capture and understand the syntactic and semantic structures of language, enabling them to generate coherent and contextually relevant text.

The widespread adoption of LLMs highlights the critical need to evaluate their robustness, as these models can be attacked to generate harmful or misleading outputs. This can have serious consequences, especially in critical domains like healthcare or finance. Therefore, it is crucial to evaluate these models carefully to ensure their reliability and safety, identifying potential vulnerabilities before they can be exploited in real-world scenarios~\citep{zhu2023promptbench, wang2021adversarial, wang2023robustness}. 
% Improving the robustness of language models against such attacks is an ongoing area of research. One common approach is adversarial training, where models are exposed to adversarial examples during training, enabling them to recognize and resist such manipulations~\citep{liu2020adversarial, altinisik-etal-2023-impact}.

%Due to their impressive capabilities, language models have become integral to a wide range of applications, and millions of people rely on them. However, this widespread usage also poses significant challenges, as it has become evident that these models can be vulnerable to adversarial attacks. 

\section{Methodology}
\begin{figure*}[t]
    \centering
        \centering
        \includegraphics[width=\linewidth]{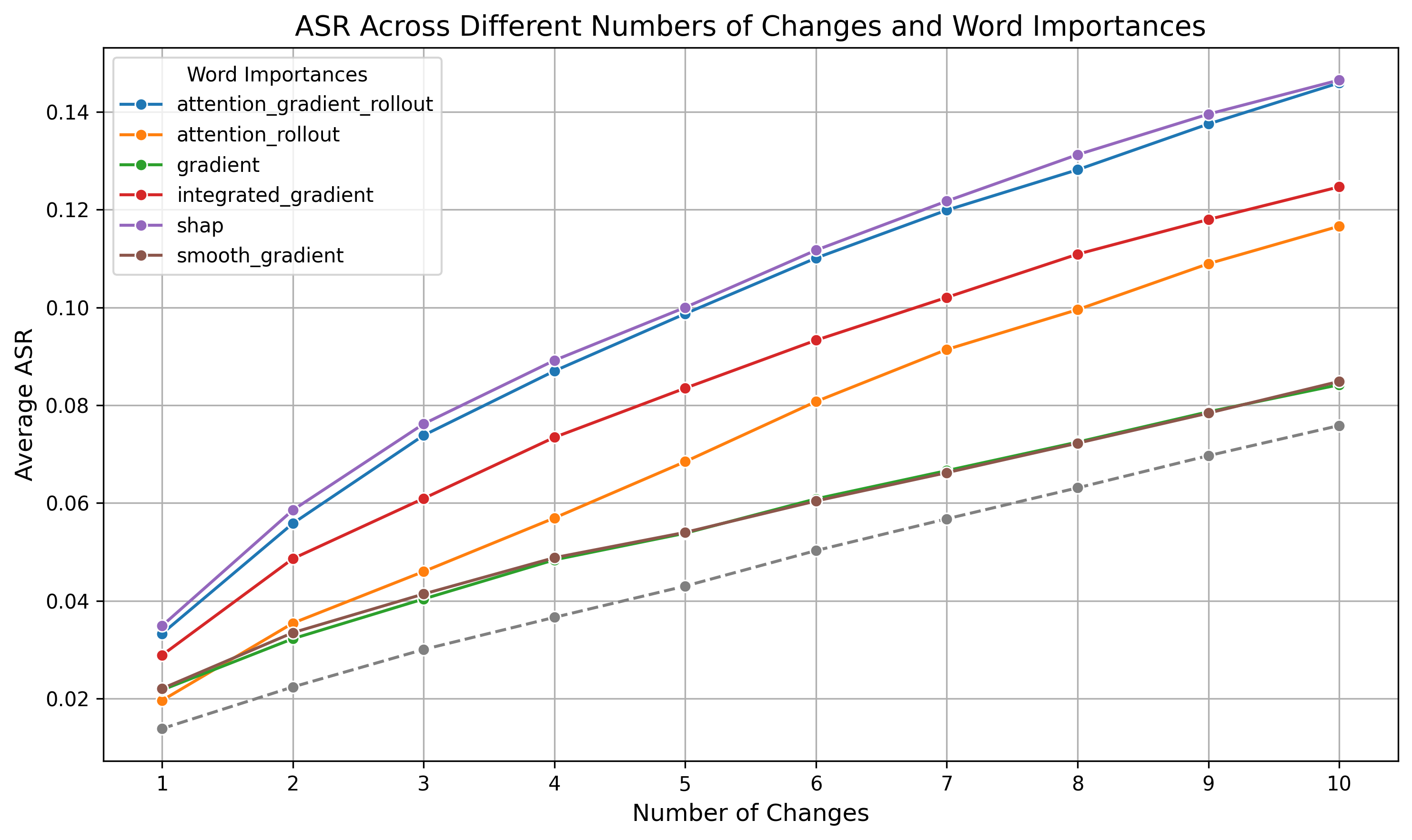}
    \caption{Relation of ASR on proxy models and number of perturbed words for different attribution methods. The dotted lines show ASR when selecting random words for perturbation instead of words based on word importance.}
    \label{fig:ASR_for_multiple_changes_LMs_importance}
\end{figure*}
\subsection{Proxy Model}
To identify the most important words, we could rank those words manually, but manually selecting the most important words is unfeasible. Thus, we used attribution methods to select the most important words. Most attribution methods need a model to calculate importance. Pre-trained LLMs could be used as such a model. Unfortunately, these models take a lot of VRAM on their own, and calculating gradient-based methods for LLMs can be very time-consuming and, for some methods, even impossible due to VRAM constraints.

We train a small proxy model to address this issue, which we later use to calculate word importance. If the model performs well on a dataset, it can be considered a good proxy for calculating word importance. Due to attribution methods highlighting what was important for the model's predictions, it's crucial to use a model with high performance for this step.
\subsection{Word Importance}
To identify the importance of individual words, we first calculate token attributions using gradient-based and perturbation-based methods. For example, Grad x Input attribution for an input sequence $X$ and output $y$ is calculated as follows
\begin{equation}
    \text{Grad x Input}(X, y) = \nabla_y(X_{emb}) \cdot X_{emb},
\end{equation}
where $X_{emb}\in \mathcal{R}^{n,d}$ are word embeddings of $n$ input tokens and $d$ is the embedding size.

However, due to the nature of tokenization, tokens may not always correspond to complete words. To address this issue, we group the tokens that form each word and aggregate their attributions using a simple mean, as shown in the following equation of importance ($\mathcal{I}$):
\begin{equation}
    \mathcal{I}(W)=\frac{1}{|S(W)|} \sum_{t\in S(W)}a(t),
\end{equation}
where $S$ is a function that splits a word $W$ into its subtokens, $|S(W)|$ denotes the number of subtokens in the word, and $a(t)$ represents the attribution value of a subtoken $t$ obtained from the chosen attribution method. This approach allows us to determine the importance of each word independently of the tokenization process, providing a more interpretable and coherent measure of word-level importance. 

We exclude tokens representing punctuation marks to ensure a more faithful aggregation of word importance. Without excluding these tokens, we would additionally consider the importance of punctuation when calculating word importance. This could lead to word importance scores that are less faithful to the semantic content of the text. By focusing on the semantic content of the text rather than the syntactic structure, we obtain a clearer picture of the importance of individual words for prediction.

\subsection{Importance-based Perturbation}
We performed perturbations that we split into two distinct levels: Character Level and Word Level.

\paragraph{Character-level Perturbations}
At the character level, we explored various techniques designed to introduce typographical errors. The specific perturbations include:
\begin{itemize}
    \item  \textbf{Keyboard errors} (\textbf{Key}): Simulating common typing mistakes resulting from adjacent key presses.  
    \item \textbf{Optical character recognition (OCR) errors}: Introducing errors caused by graphical similarities between characters, typically observed in OCR systems.
    \item \textbf{Random character insertion} (\textbf{R-Ins}): Inserting random characters within words.
    \item \textbf{Character deletion}  (\textbf{R-Del}): Removing characters from words.
    \item \textbf{Character substitution} (\textbf{R-Sub}): Replacing characters with random ones.
    \item \textbf{Characters swapping}  (\textbf{R-Sw}): Reordering adjacent characters.
    \item \textbf{Diacritical errors}  (\textbf{Dia}): Omitting diacritical marks, which can drastically change the meaning of words. For example, altering "kąt" (\textit{angle}) to "kat" (\textit{executioner}) or "język" (\textit{language}) to "jeżyk" (\textit{little hedgehog}) demonstrates how such errors can significantly affect word interpretation in languages like Polish.
 \end{itemize}

Implementing character-level perturbations in other languages necessitates updating the dictionary with diacritical errors common in the language of interest. The other perturbations do not require alteration.
\paragraph{Word-level Perturbations}
At the word level, we applied several perturbation methods to assess their impact on model performance: 
\begin{itemize}
    \item  \textbf{Random spacing}  (\textbf{Split}): Inserting spaces at random positions within words to disrupt their conventional structure. 
    \item \textbf{Orthographic errors}  (\textbf{Ort}): Common in Polish, these errors typically involve the substitution of distinctive characters. For example, replacing "rz" with "ż" or "h" with "ch". Certain spelling mistakes can lead to significant changes in word meaning, %as illustrated by the difference between 
    e.g., "morze" (\textit{sea}) and "może" (\textit{maybe}).

    \item \textbf{Relations based on Słowosieć }~\citep{maziarz2014slowosiec} (\textbf{Rel}): The Polish counterpart to WordNet, which provides not only synonyms but also hypernyms, meronyms, and other lexical relations. This resource also accounts for adjectives across different grammatical genders. For instance, substituting "sympatyczny" with "miły" (\textit{nice} to \textit{pleasant}, masculine) or "sympatyczna" with "miła" (%\textit{nice} to \textit{pleasant}, 
    feminine).
\end{itemize}
 
To extend word-level perturbations to other languages, one must update the orthographic errors specific to the language of interest. Additionally, performing changes based on word relations necessitates access to a word relation network for that language.

It is important to note that not all perturbations transfer easily to other languages. We believe many language-specific perturbations will not work for Polish. Therefore, carefully adding and removing perturbations into our framework is crucial for evaluating models accurately with language-specific variations.
% \paragraph{Sentence-level perturbations}
% To modify sentences while preserving their original meaning, we used \textbf{paraphrasing} with LLaMA 3.1 8B~\citep{dubey2024llama3herdmodels}. 

\begin{table*}[!ht]
\small  
\renewcommand{\arraystretch}{1.5} 
\caption{ASR of perturbation methods against proxy models across all datasets aggregated over attribution methods and number of words changed.}
\centering
\resizebox{\textwidth}{!}{
\begin{tabular}{l|l|rrrrrrrrrr|r}
\toprule
\textbf{} & \textbf{Data} & Diac & Key & OCR & Ort & R-Del & R-Ins & R-Sub & R-Sw & Rel & Split & \textbf{Avg} \\
\midrule
\multirow{5}{*}{\rotatebox{90}{PolBERT}} & AC & 0.01 & 0.11 & 0.13 & 0.02 & 0.08 & 0.11 & 0.11 & 0.09 & 0.02 & 0.07 & 0.08 \\
 & AR & 0.03 & 0.22 & 0.23 & 0.06 & 0.24 & 0.23 & 0.23 & 0.24 & 0.12 & 0.24 & 0.18 \\
 & CBD & 0.01 & 0.04 & 0.04 & 0.01 & 0.04 & 0.03 & 0.03 & 0.05 & 0.01 & 0.05 & 0.03 \\
 & P-I & 0.00 & 0.04 & 0.04 & 0.01 & 0.04 & 0.04 & 0.05 & 0.04 & 0.02 & 0.04 & 0.03 \\
 & P-O & 0.01 & 0.16 & 0.19 & 0.02 & 0.14 & 0.15 & 0.16 & 0.14 & 0.04 & 0.15 & 0.12 \\
\midrule
\multirow{5}{*}{\rotatebox{90}{HerBERT}} & AC & 0.01 & 0.11 & 0.12 & 0.02 & 0.04 & 0.09 & 0.11 & 0.05 & 0.02 & 0.05 & 0.06 \\
 & AR & 0.02 & 0.14 & 0.14 & 0.04 & 0.12 & 0.12 & 0.14 & 0.12 & 0.07 & 0.12 & 0.10 \\
 & CBD & - & - & - & - & - & - & - & - & - & - & - \\
 & P-I & 0.00 & 0.03 & 0.03 & 0.00 & 0.02 & 0.03 & 0.03 & 0.02 & 0.01 & 0.02 & 0.02 \\
 & P-O & 0.00 & 0.13 & 0.12 & 0.02 & 0.11 & 0.11 & 0.13 & 0.12 & 0.04 & 0.09 & 0.09 \\
\midrule
\multirow{5}{*}{\rotatebox{90}{RoBERTa}} & AC & 0.04 & 0.21 & 0.24 & 0.06 & 0.16 & 0.20 & 0.20 & 0.17 & 0.05 & 0.14 & 0.15 \\
 & AR & 0.03 & 0.22 & 0.23 & 0.05 & 0.21 & 0.21 & 0.22 & 0.21 & 0.10 & 0.20 & 0.17 \\
 & CBD & 0.00 & 0.02 & 0.02 & 0.01 & 0.02 & 0.02 & 0.02 & 0.03 & 0.01 & 0.03 & 0.02 \\
 & P-I & 0.00 & 0.04 & 0.05 & 0.01 & 0.04 & 0.04 & 0.04 & 0.05 & 0.02 & 0.04 & 0.03 \\
 & P-O & 0.01 & 0.12 & 0.13 & 0.03 & 0.12 & 0.11 & 0.12 & 0.13 & 0.04 & 0.11 & 0.09 \\
\cline{1-13}
\bottomrule
\end{tabular}
}
\label{table:results_aug}
\end{table*}
\section{Experiments}
\begin{table}[!ht]
    \centering
    \caption{Performance of Proxy Models on test sets of all datasets.}
    \resizebox{0.45\textwidth}{!}{
    \begin{tabular}{l|l|rrr}
    \toprule
         \textbf{Model} &  \textbf{Data} & \textbf{AUROC} & \textbf{F1} & \textbf{ACC} \\
        \midrule
        \multirow[t]{5}{*}{PolBERT} & AC & 0.922 & 0.828 & 0.843 \\
         & AR & 0.836 & 0.480 & 0.580 \\
         & CBD & 0.837 & 0.684 & 0.893 \\
         & P-I & 0.969 & 0.816 & 0.856 \\
         & P-O & 0.896 & 0.537 & 0.678 \\
        \midrule
        \multirow[t]{5}{*}{HerBERT} & AC & 0.926 & 0.836 & 0.851 \\
         & AR & 0.887 & 0.572 & 0.653 \\
         & CBD & 0.547 & 0.464 & 0.866 \\
         & P-I & 0.970 & 0.856 & 0.892 \\
         & P-O & 0.910 & 0.529 & 0.711 \\
        \midrule
        \multirow[t]{5}{*}{RoBERTa} & AC & 0.926 & 0.832 & 0.848 \\
         & AR & 0.863 & 0.542 & 0.624 \\
         & CBD & 0.879 & 0.660 & 0.887 \\
         & P-I & 0.972 & 0.860 & 0.886 \\
         & P-O & 0.904 & 0.529 & 0.715 \\
        \cline{1-5}
        \bottomrule
    \end{tabular}}
    \label{tab:lm_perf}
\end{table}

\subsection{Datasets}

In the experiments, we used datasets from the KLEJ benchmark~\citep{rybak2020klej}. KLEJ is the Polish equivalent of the English GLUE benchmark for text analysis.

\noindent\textbf{AC - Polish Abusive Clauses Dataset}~\citep{NEURIPS2022_890b206e} is used for detecting abusive clauses in legal agreements to protect consumers from unfair terms. It contains two classes: \textit{abusive clause} and \textit{ correct agreement statement}.

\noindent\textbf{AR - Allegro Reviews}~\citep{rybak2020klej} consists of sentiment-labeled reviews from the Polish e-commerce platform Allegro. Each opinion is assigned a value from 1 to 5, representing the sentiment score.

\noindent\textbf{CBD - Cyberbullying Detection}~\citep{ptaszynski2019results} dataset contains Twitter messages and is used to predict whether a given message contains cyberbullying or harmful content.

\noindent\textbf{P-I \& P-O - PolEmo2.0}~\citep{kocon-etal-2019-multi} is a collection of online reviews from the medicine and hotel domains. The task is to predict the sentiment of a review. Two separate test sets allow for in-domain (medicine and hotels) and out-of-domain (products and university) evaluation.

These datasets provide a diverse range of NLP tasks, including sentiment analysis, abusive content detection, and domain-specific text classification, which are crucial for evaluating the performance of models in various Polish language understanding tasks. Details about the sizes of train and test splits are available in the Appendix~\ref{sec:appendix_model_performance}.

\subsection{Models}

In our experiments, we used transformer-based classifiers HerBERT~\citep{mroczkowski-etal-2021-herbert}, PolBERT~\citep{Kleczek2020polbert}, and Polish RoBERTa~\citep{dadas2024assessing} as proxy models. These models were fine-tuned on the training set of each analyzed dataset for the specific task. The training process involved 20 epochs on an A100 (40GB) GPU with early stopping, and the batch size was set to 16 (8 for Allegro reviews), for models performance, see Table \ref{tab:lm_perf}. After training, these models were additionally evaluated for their robustness against perturbations using the test sets. We observed that the HerBERT model failed to train on the CBD dataset, which led to it predicting the same label for all examples. Thus, we omitted using it for further experiments. 

For LLMs, we used open-source generative models capable of processing the Polish language, such as Bielik~\citep{ociepa2024bielik}, Mistral-7B-Instruct~\citep{jiang2023mistral}, and Llama-3.1-8 B~\citep {dubey2024llama3herdmodels}. We applied a zero-shot learning approach for these models, where the models were queried using both original and modified prompts.

As a final proxy model for the word importance, we used a model based on the polish-roberta-base-v2 classifier, which showed the highest performance for all datasets, and the SHAP method as an attribution method, due to SHAP having the highest success rate of the perturbations for proxy models.

\begin{table*}[!ht]
\small  
\renewcommand{\arraystretch}{1.5} 
\caption{ASR of perturbation methods against smaller models across all datasets aggregated over the number of words changed. Perturbations were generated using SHAP word importance scores and a RoBERTa classifier. Robustness values above the threshold are highlighted in red, with more intense red indicating lower robustness.  Values below the threshold are highlighted in green, with more intense green indicating higher robustness.}
\centering
\resizebox{\textwidth}{!}{
\begin{tabular}{l|l|cccccccccc}
\toprule
\textbf{} & \textbf{Data} & Diac & Key & OCR & Ort & R-Del & R-Ins & R-Sub & R-Sw & Rel & Split% & \textbf{Avg}
\\
\midrule
\multirow{5}{*}{\rotatebox{90}{Bielik v1}} & AC & \cgr{0.145} & \cgr{0.285} & \cgr{0.267} & \cgr{0.170} & \cgr{0.261} & \cgr{0.261} & \cgr{0.288} & \cgr{0.283} & \cgr{0.182} & \cgr{0.224} %&  \cgr{0.237}
 \\
& AR & \cgr{0.082} & \cgr{0.196} & \cgr{0.200} & \cgr{0.083} & \cgr{0.176} & \cgr{0.170} & \cgr{0.191} & \cgr{0.173} & \cgr{0.145} & \cgr{0.151} %& \cgr{0.197}
 \\
 & CBD & \cgr{0.087} & \cgr{0.452} & \cgr{0.488} & \cgr{0.156} & \cgr{0.318} & \cgr{0.441} & \cgr{0.477} & \cgr{0.360} & \cgr{0.173} & \cgr{0.383} %& \cgr{0.245}
  \\
 & P-I & \cgr{0.134} & \cgr{0.302} & \cgr{0.308} & \cgr{0.167} & \cgr{0.264} & \cgr{0.263} & \cgr{0.286} & \cgr{0.265} & \cgr{0.242}
 & \cgr{0.273} %& \cgr{0.292}  
 \\
 & P-O & \cgr{0.149} & \cgr{0.432} & \cgr{0.417} & \cgr{0.174} & \cgr{0.380} & \cgr{0.427} & \cgr{0.414} & \cgr{0.389}  & \cgr{0.307}
 & \cgr{0.361} %& \cgr{0.298}
\\
\midrule
\multirow{5}{*}{\rotatebox{90}{Mistral 7B}} & AC & \cgr{0.010} & \cgr{0.076} & \cgr{0.087} & \cgr{0.014} & \cgr{0.041} & \cgr{0.068} & \cgr{0.076} & \cgr{0.056} & \cgr{0.022}	& \cgr{0.042} %& \cgr{0.076}
  \\
 & AR & \cgr{0.027} & \cgr{0.175} & \cgr{0.202} & \cgr{0.048} & \cgr{0.156} & \cgr{0.141} & \cgr{0.177} & \cgr{0.156} & \cgr{0.103}	& \cgr{0.146} %& \cgr{0.091}
  \\
 & CBD & \cgr{0.013} & \cgr{0.348} & \cgr{0.368} & \cgr{0.029} & \cgr{0.156} & \cgr{0.335} & \cgr{0.355} & \cgr{0.213} & \cgr{0.056}	& \cgr{0.214} %& \cgr{0.171}
  \\
 & P-I & \cgr{0.003} & \cgr{0.028} & \cgr{0.029} & \cgr{0.010} & \cgr{0.027} & \cgr{0.032} & \cgr{0.032} & \cgr{0.026} & \cgr{0.016} &	\cgr{0.025} %& \cgr{0.116}
  \\
 & P-O & \cgr{0.015} & \cgr{0.075} & \cgr{0.073} & \cgr{0.024} & \cgr{0.074} & \cgr{0.058} & \cgr{0.078} & \cgr{0.061} & \cgr{0.034} &	\cgr{0.065} %& \cgr{0.039}
 \\
\midrule
\multirow{5}{*}{\rotatebox{90}{Llama3 8B}} & AC & \cgr{0.219} & \cgr{0.333} & \cgr{0.334} & \cgr{0.241} & \cgr{0.314} & \cgr{0.332} & \cgr{0.337} & \cgr{0.325} &  \cgr{0.239}
 & \cgr{0.302} %& \cgr{0.321}
 \\
 & AR & \cgr{0.075} & \cgr{0.346} & \cgr{0.346} & \cgr{0.166} & \cgr{0.345} & \cgr{0.354} & \cgr{0.329} & \cgr{0.296} &   \cgr{0.221}	& \cgr{0.261} %& \cgr{0.286}  
 \\
 & CBD & \cgr{0.041} & \cgr{0.148} & \cgr{0.232} & \cgr{0.058} & \cgr{0.102} & \cgr{0.159} & \cgr{0.157} & \cgr{0.127} &  \cgr{0.062}	& \cgr{0.093} %& \cgr{0.196}  
 \\
 & P-I & \cgr{0.004} & \cgr{0.051} & \cgr{0.044} & \cgr{0.005} & \cgr{0.035} & \cgr{0.043} & \cgr{0.052} & \cgr{0.037} & \cgr{0.013} &	\cgr{0.018} %& \cgr{0.074}
 \\
 & P-O & \cgr{0.031} & \cgr{0.137} & \cgr{0.146} & \cgr{0.042} & \cgr{0.111} & \cgr{0.131} & \cgr{0.136} & \cgr{0.118} &  \cgr{0.070}	& \cgr{0.098} %& \cgr{0.066}
 \\

\midrule
\bottomrule
\end{tabular}
}
\label{table:results_aug_llm}
\end{table*}

\subsection{Metric}
To evaluate the effectiveness of perturbations, we utilized the \textbf{Attack Success Rate (ASR)} metric. This metric serves as a standard measure for assessing the performance of adversarial attacks by calculating the proportion of cases where the attack achieves its intended goal. A higher ASR indicates a more effective attack, whereas a lower ASR suggests greater resistance of the model to adversarial manipulations. Formally, ASR can be defined as:
\begin{equation}
    ASR = \sum_{(x,y) \in \mathcal{D} }\frac{\mathbb{1}[f( \mathcal{A}(x))\neq y]\cdot \mathbb{1}[f(x)=y]}{\sum_{(x',y') \in \mathcal{D} }\mathbb{1}[f(x')=y']},
\end{equation}
 
\noindent where $\mathcal{D}$ denotes the dataset of input samples $x$ and their corresponding true labels $y$, $f(x)$ represents the model's prediction for a given input $x$, $\mathcal{A}(x)$ is the adversarially perturbated version of the input $x$ generated by the perturbation $\mathcal{A}$ and $\mathbb{1}[]$ is the indicator. % function.

% \begin{table*}[!ht]
% \small  
% \renewcommand{\arraystretch}{1.5} 
% \centering
% \caption{ASR of paraphrasing attack using LLaMA-3.1-8B model across all datasets.}
% \begin{tabular}{l|cccccc}
% \toprule
% \multicolumn{1}{c|}{\multirow{2}{*}{\textbf{Data}}} & \multicolumn{6}{c}{\textbf{Model}} \\
% \cmidrule{2-7}
%  & PolBERT & HerBERT & RoBERTa & Bielik-7B & Mistral-7B & Llama-3.1-8B \\
% \midrule
% AC & 0.13 & 0.11 & 0.13 & 0.24 & 0.04  & 0.25  \\
% AR & 0.31 & 0.24 & 0.25 & 0.33 &  0.21& 0.34 \\
% CBD & 0.04 & - & 0.03 & 0.23 & 0.11 & 0.06 \\
% P-I & 0.09 & 0.04 & 0.07 & 0.39 & 0.03 & 0.05  \\
% P-O & 0.35 & 0.18 & 0.15 & 0.48 & 0.07 & 0.08 \\ \cline{1-7}
% \bottomrule
% \end{tabular}
% \label{table:results_llm}
% \end{table*}

\section{Results}
\begin{figure*}[!t]
    \centering
        \centering
        \includegraphics[width=\linewidth]{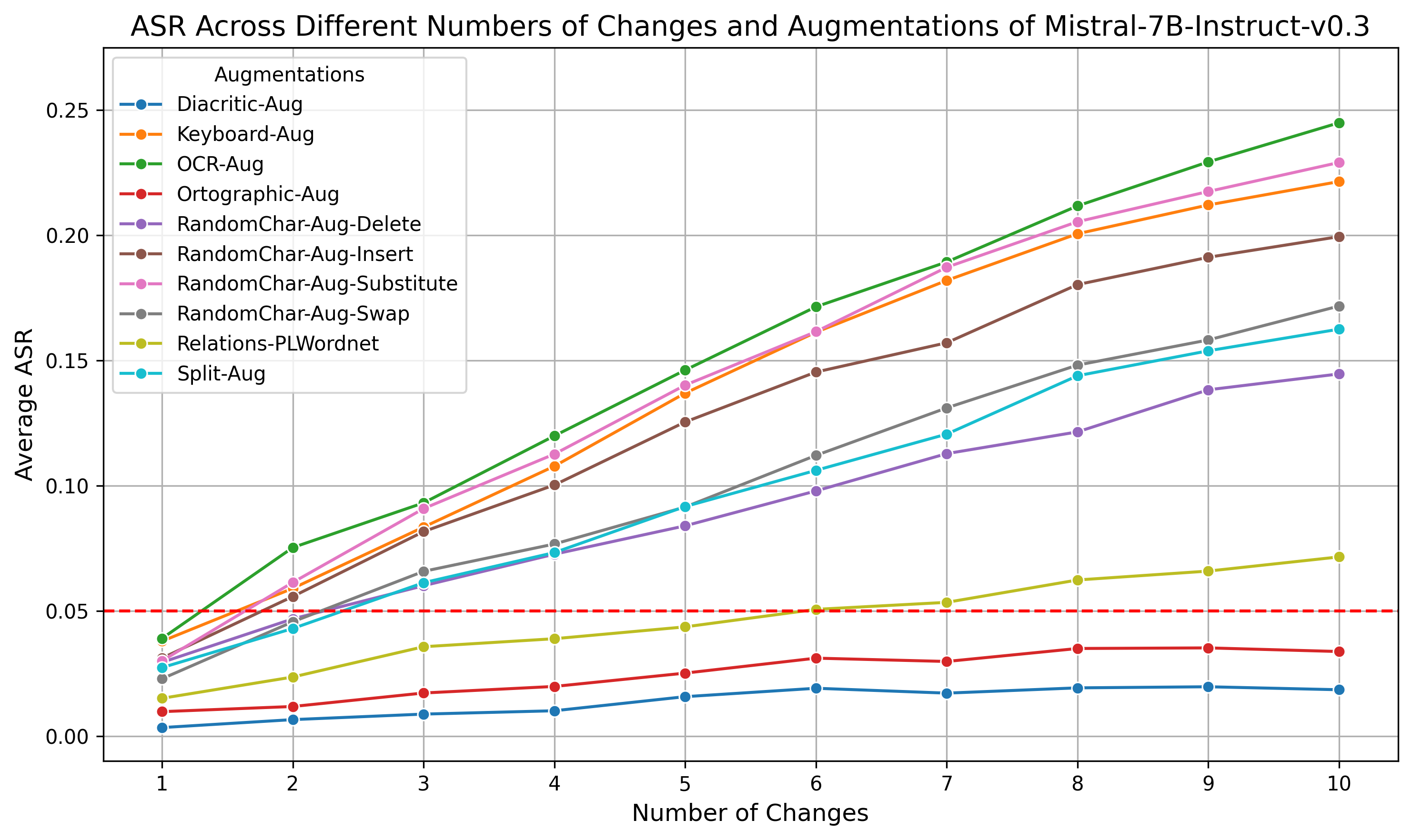}
    \caption{Relation of ASR on Mistral and the number of perturbed words for different attribution methods. The dotted line indicates the robustness cutoff above which the model is considered non-robust to simple perturbations.}
    \label{fig:ASR_for_multiple_changes_mistral}
\end{figure*}
\subsection{Influence of Perturbation Type}
Results in Table~\ref{table:results_aug} highlight that our LMs are fooled into changing their prediction when faced with simple character perturbations, such as OCR, R-Del, R-Ins, R-sub, R-Sw, and Split. These models haven't seen such errors in training data, potentially making them perform poorly on examples with such perturbations. %Interestingly, both classifiers and LLMs demonstrate a lack of robustness to paraphrased texts, as their predictions are often inconsistent when faced with rephrased versions of the same original content (see Table~\ref{table:results_llm}).

In Table~\ref{table:results_aug_llm}, we can observe that even though LLMs were trained on data from the internet, which should have examples with character-level errors, those models can be easily misguided by such perturbations. This should raise the attention of the developers to fix this issue. These are simple classification tasks, but if models make more errors when faced with such perturbations on those tasks, there is a high risk that those types of perturbations could be used to deceive models into generating harmful content.
\subsection{Influence of Attribution Method}
Based on the results in Figure~\ref{fig:ASR_for_multiple_changes_LMs_importance}, we can observe that SHAP gives the highest ASR compared to other methods for all numbers of words changed. %On the other hand, we can observe that 
Vanilla Gradient and SmoothGrad have the worst performance in terms of ASR and perform very similarly to selecting random words. This hints that it will be beneficial to use SHAP to extract word importance when selecting words that should be perturbed, in our case. 
\subsection{Robustness of LLMs}
Figure~\ref{fig:ASR_for_multiple_changes_mistral} and Table~\ref{table:results_aug_llm} present an example usage of the proposed framework for the Mistral model. The red line denotes the threshold ($0.05$) equivalent to the model being fooled in $5\%$ of examples with these perturbations, and we decide that when the model is fooled so often, then it could be the case that it has vulnerability related to such perturbations. 

The analysis reveals that while the model effectively handles diacritic and orthographic perturbations, it is vulnerable to other modifications, such as inserting additional characters at the word level. This finding is particularly significant from an AI safety perspective, as it suggests that although the model may resist harmful prompts in their standard form, subtle alterations, such as introducing spaces within key terms, could increase the likelihood of generating harmful responses.

\section{Conclusions}
In this work, we introduced a framework to assess the robustness of LLMs by utilizing perturbations and word importance calculated with proxy models. We validated this framework in Polish by curating a set of perturbed datasets that can be used to assess LLMs' robustness in Polish.

To evaluate robustness to perturbations, we perform the following steps:
\begin{enumerate}
    \item \textbf{Create Proxy Model:} Train a small model on the target dataset.
    \item \textbf{Rank:} Calculate and aggregate attribution scores based on the proxy model to create a word-importance ranking.
    \item \textbf{Perturb:} Perturb the most important words according to the ranking.
    \item \textbf{Evaluate:} Evaluate the target LLMs on the original and perturbed datasets to assess their robustness.
\end{enumerate}

Compatibility with various perturbation methods and tasks is a key feature of this framework, assuming attribution calculation is feasible. Using perturbed datasets prepared using our framework is particularly advantageous in LLMs development. By carefully preparing datasets with specified perturbations, developers can effectively assess and control the robustness of the models they create. Should a perturbation lead to an LLM's failure on a classification task, it underscores a vulnerability that warrants attention, given the potential for such perturbations to negatively influence model performance in safety-critical generative tasks, such as the generation of instructions for prohibited activities.

Our investigation utilizing the framework for the Polish language revealed that such perturbations, which can be implemented with minimal resources, are unexpectedly effective in deceiving LLMs. We highlight the perturbations that proved most successful in fooling these models, indicating areas requiring attention.
\section{Limitations}

Our word importance relies on attribution methods, which may not faithfully reflect the model's internal reasoning processes. On the other hand, the choice of model used for generating attributions can also influence whether the words selected are really important, whether these reflect genuine semantic importance, or merely artifacts specific to the proxy model's training. 

The analysis relies on heuristic perturbations derived from word importance rankings. This approach does not encompass the full spectrum of potential attacks against LLMs, such as those generated through adversarial optimization techniques.

Another consideration is that the process of perturbing important words could make key segments of the text unintelligible, even from a human perspective. Such semantic degradation may confound the assessment, making isolating model robustness issues from the effects of corrupted input challenging.

Moreover, relying on proxy models for generating word importance scores is an approximation. Importance rankings derived directly from the target LLMs could potentially differ and offer a more precise basis for identifying words whose perturbation maximally impacts the LLM's performance.

%%%%%%%%%%%%%%%%%%%%%%%%%%%%%%%%%%%%%%%%%%%%%%%%%%%%%%%%%%%%%%%%%%%%%%%%

%%% Use this environment to include acknowledgements (optional).
%%% This will be omitted in doubleblind mode.

% \begin{ack}
% By using the \texttt{ack} environment to insert your (optional) 
% acknowledgements, you can ensure that the text is suppressed whenever 
% you use the \texttt{doubleblind} option. In the final version, 
% acknowledgements may be included on the extra page intended for references.
% \end{ack}

%%%%%%%%%%%%%%%%%%%%%%%%%%%%%%%%%%%%%%%%%%%%%%%%%%%%%%%%%%%%%%%%%%%%%%%%

%%% Use this command to include your bibliography file.

% \bibliography{mybibfile}
\bibliographystyle{unsrt}  
\bibliography{references}

\appendix
\clearpage

\label{sec:appendix}

\section{Models and Datasets}
\label{sec:appendix_model_performance}
Information about dataset sizes is available in Table~\ref{tab:datasets_statistics} contains LLMs performance on original test sets.
\begin{table}[h]
    \centering
    \caption{Sizes of train and tests splits of used datasets.}
    \resizebox{0.45\textwidth}{!}{
    \begin{tabular}{l|rr}
        \toprule
        \textbf{Dataset} & \textbf{Train} & \textbf{Test} \\
        \midrule
        Polish Abusive Clauses & 4284 & 3453 \\
        Allegro Reviews & 9577 & 1006 \\
        Cyberbullying & 10041 & 1000 \\
        PolEmo2.0 In & 5783 & 722 \\
        PolEmo2.0 Out & 5783 & 494 \\
        \cline{1-3}\bottomrule
    \end{tabular}
    }
    \label{tab:datasets_statistics}
\end{table}

\begin{table}[h]
    \centering
    \caption{Performance of LLMs on test sets.}
    \resizebox{0.45\textwidth}{!}{
    \begin{tabular}{l|l|rr}
    \toprule
         \textbf{Model} &  \textbf{Data} &  \textbf{F1} & \textbf{ACC} \\
        \midrule
        \multirow[t]{5}{*}{Bielik-7B} & AC &0.54 & 0.54  \\
         & AR & 0.53 & 0.50  \\
         & CBD & 0.71 & 0.65   \\
         & P-I & 0.42 & 0.40\\
         & P-O & 0.43 & 0.33 \\
        \midrule
        \multirow[t]{5}{*}{Mistral-7B} & AC & 0.60   & 0.54 \\
         & AR & 0.50 & 0.49  \\
         & CBD &  0.83& 0.82   \\
         & P-I & 0.58 & 0.68  \\
         & P-O & 0.57 & 0.66  \\
        \midrule
        \multirow[t]{5}{*}{Llama-3.1-8B} & AC & 0.42  &  0.45 \\
         & AR & 0.50  & 0.47   \\
         & CBD &  0.85& 0.87   \\
         & P-I & 0.60 & 0.69   \\
         & P-O & 0.61 & 0.67  \\
        \midrule
        \bottomrule
    \end{tabular}
    }
    \label{tab:llm_perf}
\end{table}
% \begin{table}[h]
%     \centering
%     \caption{Performance of LMs on test sets.}
%     \begin{tabular}{l|l|rrr}
%     \toprule
%          \textbf{Model} &  \textbf{Data} & \textbf{ROC} & \textbf{F1} & \textbf{ACC} \\
%         \midrule
%         \multirow[t]{5}{*}{PolBERT} & AC & 0.922 & 0.828 & 0.843 \\
%          & AR & 0.836 & 0.480 & 0.580 \\
%          & CBD & 0.837 & 0.684 & 0.893 \\
%          & P-I & 0.969 & 0.816 & 0.856 \\
%          & P-O & 0.896 & 0.537 & 0.678 \\
%         \cline{1-5}
%         \multirow[t]{5}{*}{HerBERT} & AC & 0.926 & 0.836 & 0.851 \\
%          & AR & 0.887 & 0.572 & 0.653 \\
%          & CBD & 0.547 & 0.464 & 0.866 \\
%          & P-I & 0.970 & 0.856 & 0.892 \\
%          & P-O & 0.910 & 0.529 & 0.711 \\
%         \cline{1-5}
%         \multirow[t]{5}{*}{RoBERTa} & AC & 0.926 & 0.832 & 0.848 \\
%          & AR & 0.863 & 0.542 & 0.624 \\
%          & CBD & 0.879 & 0.660 & 0.887 \\
%          & P-I & 0.972 & 0.860 & 0.886 \\
%          & P-O & 0.904 & 0.529 & 0.715 \\
%         \cline{1-5}
%         \bottomrule
%     \end{tabular}
%     \label{tab:lm_perf}
% \end{table}
\section{Perturbations}
\subsection{Character-level}
All character-level perturbations randomly select the number of characters to change between $1$ and $\min(len(word)\cdot0.15, 4)$. It ensures that character-level changes make perturbed words easily understandable to a human.
\subsection{Word-level}
Word-level perturbations always try to modify the provided word, but some methods can be unsuccessful.
\subsection{Word importance}
To calculate SmoothGrad and Integrated Gradients attributions, we used 50 steps. For SHAP attributions, we use default arguments provided by SHAP~\citep{scott2017unified} library.

\section{Perturbation attacks success rate}
\begin{table}[ht!]
\small  
\renewcommand{\arraystretch}{1.5}
\small
\centering
\caption{ASR of attribution methods against smaller models across all datasets.}
\resizebox{0.5\textwidth}{!}{
\begin{tabular}{l|l|rrrrrr|r}
\toprule
\textbf{Model} & \textbf{Data} & AGR & AR & G & IG & SH & SG & \textbf{Avg} \\
\midrule
\multirow{5}{*}{PolBERT} & AC & 0.08 & 0.09 & 0.07 & 0.07 & 0.08 & 0.07 & 0.08 \\
 & AR & 0.28 & 0.16 & 0.12 & 0.20 & 0.22 & 0.12 & 0.18 \\
 & CBD & 0.03 & 0.03 & 0.03 & 0.03 & 0.03 & 0.03 & 0.03 \\
 & P-I & 0.07 & 0.03 & 0.02 & 0.03 & 0.05 & 0.01 & 0.04 \\
 & P-O & 0.18 & 0.12 & 0.07 & 0.11 & 0.15 & 0.07 & 0.12 \\
\midrule
\multirow{5}{*}{HerBERT} & AC & 0.06 & 0.09 & 0.05 & 0.05 & 0.07 & 0.05 & 0.06 \\
 & AR & 0.11 & 0.09 & 0.08 & 0.10 & 0.14 & 0.08 & 0.10 \\
 & CBD & - & - & - & - & - & - & - \\
 & P-I & 0.03 & 0.02 & 0.02 & 0.02 & 0.03 & 0.02 & 0.02 \\
 & P-O & 0.12 & 0.08 & 0.06 & 0.10 & 0.11 & 0.07 & 0.09 \\
\midrule
\multirow{5}{*}{RoBERTa} & AC & 0.15 & 0.15 & 0.13 & 0.16 & 0.15 & 0.13 & 0.14 \\
 & AR & 0.21 & 0.14 & 0.10 & 0.21 & 0.25 & 0.10 & 0.17 \\
 & CBD & 0.02 & 0.02 & 0.01 & 0.02 & 0.03 & 0.01 & 0.02 \\
 & P-I & 0.04 & 0.02 & 0.02 & 0.04 & 0.06 & 0.02 & 0.03 \\
 & P-O & 0.11 & 0.07 & 0.06 & 0.11 & 0.14 & 0.07 & 0.09 \\
\midrule
\bottomrule
\end{tabular}
}
\label{table:results_word_importance}
\end{table}
ASR broken down by attribution methods is available in Table~\ref{table:results_word_importance}.

Tables \ref{tab:shap_asr_per_model},\ref{tab:gradient_asr_per_model},\ref{tab:smoothgrad_asr_per_model},\ref{tab:ig_asr_per_model},\ref{tab:attn_asr_per_model} and \ref{tab:attn_grad_asr_per_model} present the ASR broken down by model, augmentation method, and the number of words changed for classifiers.

\begin{table*}[!ht]
    \centering
        \caption{ASR for different numbers of words changed when using SHAP as attribution method, aggregated over datasets.}
    \resizebox{\textwidth}{!}{    
    \begin{tabular}{ll|rrrrrrrrrr}
    \toprule
        \textbf{Model} & \textbf{Aug} & 1 & 2 & 3 & 4 & 5 & 6 & 7 & 8 & 9 & 10  \\
        \midrule
        \multirow[t]{11}{*}{PolBERT} & Diac & 0.004 & 0.006 & 0.007 & 0.010 & 0.012 & 0.013 & 0.014 & 0.015 & 0.019 & 0.022 \\
         & Key & 0.057 & 0.091 & 0.110 & 0.124 & 0.139 & 0.152 & 0.164 & 0.173 & 0.180 & 0.188 \\
         & OCR & 0.066 & 0.101 & 0.125 & 0.142 & 0.159 & 0.170 & 0.189 & 0.195 & 0.204 & 0.205 \\
         & Ort & 0.010 & 0.015 & 0.020 & 0.027 & 0.030 & 0.035 & 0.039 & 0.042 & 0.047 & 0.051 \\
         & R-Del & 0.052 & 0.087 & 0.107 & 0.123 & 0.135 & 0.146 & 0.157 & 0.166 & 0.172 & 0.179 \\
         & R-Ins & 0.058 & 0.091 & 0.115 & 0.126 & 0.138 & 0.146 & 0.160 & 0.169 & 0.177 & 0.186 \\
         & R-Sub & 0.063 & 0.093 & 0.117 & 0.132 & 0.145 & 0.156 & 0.168 & 0.178 & 0.181 & 0.187 \\
         & R-Sw & 0.056 & 0.088 & 0.112 & 0.127 & 0.145 & 0.154 & 0.164 & 0.176 & 0.184 & 0.191 \\
         & Rel & 0.023 & 0.033 & 0.041 & 0.046 & 0.049 & 0.054 & 0.062 & 0.065 & 0.071 & 0.074 \\
         & Split & 0.053 & 0.083 & 0.107 & 0.123 & 0.134 & 0.148 & 0.159 & 0.167 & 0.175 & 0.179 \\
        \midrule
        \multirow[t]{11}{*}{HerBERT} & Diac & 0.003 & 0.004 & 0.005 & 0.006 & 0.007 & 0.008 & 0.011 & 0.012 & 0.013 & 0.016 \\
         & Key & 0.033 & 0.056 & 0.078 & 0.093 & 0.111 & 0.120 & 0.134 & 0.145 & 0.153 & 0.163 \\
         & OCR & 0.033 & 0.053 & 0.075 & 0.090 & 0.101 & 0.117 & 0.126 & 0.133 & 0.141 & 0.149 \\
         & Ort & 0.006 & 0.010 & 0.012 & 0.015 & 0.015 & 0.016 & 0.021 & 0.024 & 0.025 & 0.028 \\
         & R-Del & 0.025 & 0.042 & 0.055 & 0.068 & 0.074 & 0.085 & 0.090 & 0.103 & 0.112 & 0.121 \\
         & R-Ins & 0.027 & 0.049 & 0.064 & 0.075 & 0.086 & 0.100 & 0.111 & 0.116 & 0.124 & 0.128 \\
         & R-Sub & 0.034 & 0.056 & 0.078 & 0.096 & 0.108 & 0.121 & 0.130 & 0.144 & 0.152 & 0.165 \\
         & R-Sw & 0.027 & 0.045 & 0.061 & 0.075 & 0.083 & 0.092 & 0.100 & 0.109 & 0.117 & 0.123 \\
         & Rel & 0.011 & 0.018 & 0.023 & 0.028 & 0.034 & 0.039 & 0.043 & 0.050 & 0.055 & 0.054 \\
         & Split & 0.023 & 0.038 & 0.052 & 0.061 & 0.071 & 0.074 & 0.080 & 0.088 & 0.093 & 0.098 \\
        \midrule
        \multirow[t]{11}{*}{RoBERTa} & Diac & 0.007 & 0.008 & 0.010 & 0.011 & 0.013 & 0.018 & 0.022 & 0.026 & 0.032 & 0.037 \\
         & Key & 0.053 & 0.096 & 0.127 & 0.149 & 0.168 & 0.193 & 0.211 & 0.225 & 0.238 & 0.249 \\
         & OCR & 0.058 & 0.107 & 0.140 & 0.167 & 0.182 & 0.201 & 0.213 & 0.228 & 0.242 & 0.253 \\
         & Ort & 0.013 & 0.017 & 0.023 & 0.026 & 0.033 & 0.042 & 0.049 & 0.057 & 0.061 & 0.067 \\
         & R-Del & 0.045 & 0.082 & 0.111 & 0.132 & 0.156 & 0.180 & 0.199 & 0.213 & 0.229 & 0.241 \\
         & R-Ins & 0.043 & 0.085 & 0.116 & 0.139 & 0.155 & 0.175 & 0.188 & 0.208 & 0.223 & 0.232 \\
         & R-Sub & 0.050 & 0.094 & 0.122 & 0.146 & 0.167 & 0.186 & 0.203 & 0.220 & 0.231 & 0.246 \\
         & R-Sw & 0.048 & 0.093 & 0.123 & 0.144 & 0.161 & 0.185 & 0.200 & 0.219 & 0.239 & 0.245 \\
         & Rel & 0.021 & 0.035 & 0.044 & 0.051 & 0.056 & 0.067 & 0.072 & 0.080 & 0.091 & 0.098 \\
         & Split & 0.045 & 0.081 & 0.104 & 0.125 & 0.136 & 0.158 & 0.174 & 0.191 & 0.207 & 0.218 \\
        \midrule
        \bottomrule
    \end{tabular}
    }
    \label{tab:shap_asr_per_model}
\end{table*}

\begin{table*}[!th]
    \centering
        \caption{ASR for different numbers of words changed when using Gradient as attribution method, aggregated over datasets.}
    \resizebox{\textwidth}{!}{
    \begin{tabular}{ll|rrrrrrrrrr}
    \toprule
        \textbf{Model} & \textbf{Aug} & 1 & 2 & 3 & 4 & 5 & 6 & 7 & 8 & 9 & 10  \\
        \midrule
        \multirow[t]{11}{*}{PolBERT} & Diac & 0.005 & 0.007 & 0.008 & 0.009 & 0.010 & 0.011 & 0.013 & 0.015 & 0.018 & 0.021 \\
         & Key & 0.031 & 0.046 & 0.060 & 0.073 & 0.079 & 0.083 & 0.089 & 0.096 & 0.102 & 0.106 \\
         & OCR & 0.035 & 0.054 & 0.069 & 0.082 & 0.087 & 0.094 & 0.103 & 0.112 & 0.119 & 0.125 \\
         & Ort & 0.007 & 0.011 & 0.013 & 0.016 & 0.020 & 0.024 & 0.029 & 0.032 & 0.034 & 0.038 \\
         & R-Del & 0.027 & 0.040 & 0.054 & 0.064 & 0.072 & 0.080 & 0.084 & 0.090 & 0.096 & 0.099 \\
         & R-Ins & 0.033 & 0.046 & 0.058 & 0.067 & 0.075 & 0.080 & 0.090 & 0.092 & 0.100 & 0.107 \\
         & R-Sub & 0.032 & 0.046 & 0.060 & 0.071 & 0.078 & 0.084 & 0.090 & 0.095 & 0.103 & 0.107 \\
         & R-Sw & 0.033 & 0.050 & 0.058 & 0.073 & 0.079 & 0.085 & 0.090 & 0.096 & 0.107 & 0.110 \\
         & Rel & 0.014 & 0.022 & 0.027 & 0.029 & 0.034 & 0.036 & 0.040 & 0.043 & 0.048 & 0.052 \\
         & Split & 0.031 & 0.048 & 0.057 & 0.070 & 0.075 & 0.082 & 0.087 & 0.092 & 0.095 & 0.100 \\
        \midrule
        \multirow[t]{11}{*}{HerBERT} & Diac & 0.003 & 0.004 & 0.005 & 0.007 & 0.007 & 0.008 & 0.009 & 0.010 & 0.012 & 0.014 \\
         & Key & 0.027 & 0.036 & 0.043 & 0.054 & 0.063 & 0.070 & 0.071 & 0.076 & 0.081 & 0.088 \\
         & OCR & 0.026 & 0.039 & 0.047 & 0.057 & 0.061 & 0.068 & 0.072 & 0.078 & 0.084 & 0.087 \\
         & Ort & 0.006 & 0.008 & 0.011 & 0.012 & 0.013 & 0.013 & 0.015 & 0.017 & 0.021 & 0.023 \\
         & R-Del & 0.018 & 0.030 & 0.033 & 0.037 & 0.044 & 0.047 & 0.048 & 0.054 & 0.060 & 0.065 \\
         & R-Ins & 0.022 & 0.035 & 0.040 & 0.048 & 0.053 & 0.062 & 0.065 & 0.068 & 0.072 & 0.077 \\
         & R-Sub & 0.028 & 0.041 & 0.046 & 0.053 & 0.057 & 0.067 & 0.072 & 0.081 & 0.085 & 0.093 \\
         & R-Sw & 0.020 & 0.031 & 0.035 & 0.045 & 0.048 & 0.052 & 0.053 & 0.054 & 0.059 & 0.066 \\
         & Rel & 0.011 & 0.015 & 0.017 & 0.021 & 0.024 & 0.026 & 0.029 & 0.033 & 0.038 & 0.040 \\
         & Split & 0.020 & 0.032 & 0.035 & 0.038 & 0.041 & 0.044 & 0.049 & 0.050 & 0.057 & 0.059 \\
        \midrule
        \multirow[t]{11}{*}{RoBERTa} & Diac & 0.005 & 0.006 & 0.008 & 0.009 & 0.011 & 0.015 & 0.020 & 0.026 & 0.031 & 0.036 \\
         & Key & 0.029 & 0.044 & 0.062 & 0.075 & 0.084 & 0.096 & 0.108 & 0.119 & 0.125 & 0.135 \\
         & OCR & 0.036 & 0.056 & 0.073 & 0.085 & 0.095 & 0.109 & 0.119 & 0.125 & 0.136 & 0.143 \\
         & Ort & 0.006 & 0.009 & 0.011 & 0.013 & 0.017 & 0.023 & 0.029 & 0.034 & 0.042 & 0.048 \\
         & R-Del & 0.026 & 0.038 & 0.050 & 0.062 & 0.072 & 0.085 & 0.097 & 0.106 & 0.113 & 0.124 \\
         & R-Ins & 0.027 & 0.041 & 0.055 & 0.065 & 0.073 & 0.091 & 0.100 & 0.111 & 0.120 & 0.129 \\
         & R-Sub & 0.030 & 0.043 & 0.058 & 0.070 & 0.078 & 0.094 & 0.104 & 0.116 & 0.125 & 0.134 \\
         & R-Sw & 0.030 & 0.041 & 0.055 & 0.067 & 0.076 & 0.091 & 0.102 & 0.112 & 0.123 & 0.132 \\
         & Rel & 0.009 & 0.011 & 0.016 & 0.020 & 0.023 & 0.031 & 0.037 & 0.043 & 0.051 & 0.058 \\
         & Split & 0.025 & 0.037 & 0.049 & 0.059 & 0.066 & 0.078 & 0.086 & 0.096 & 0.102 & 0.110 \\
        \midrule
        \bottomrule
    \end{tabular}
    }
    \label{tab:gradient_asr_per_model}
\end{table*}

\begin{table*}[!ht]
    \centering
        \caption{ASR for different numbers of words changed when using SmoothGrad as attribution method, aggregated over datasets.}
    \resizebox{\textwidth}{!}{
    \begin{tabular}{ll|rrrrrrrrrr}
    \toprule
        \textbf{Model} & \textbf{Aug} & 1 & 2 & 3 & 4 & 5 & 6 & 7 & 8 & 9 & 10  \\
        \midrule
        \multirow[t]{11}{*}{PolBERT} & Diac & 0.004 & 0.006 & 0.007 & 0.008 & 0.008 & 0.010 & 0.013 & 0.015 & 0.019 & 0.021 \\
         & Key & 0.030 & 0.046 & 0.060 & 0.073 & 0.078 & 0.083 & 0.088 & 0.096 & 0.102 & 0.111 \\
         & OCR & 0.035 & 0.054 & 0.066 & 0.080 & 0.090 & 0.096 & 0.105 & 0.112 & 0.121 & 0.126 \\
         & Ort & 0.008 & 0.010 & 0.012 & 0.015 & 0.019 & 0.023 & 0.028 & 0.030 & 0.033 & 0.037 \\
         & R-Del & 0.029 & 0.041 & 0.052 & 0.066 & 0.074 & 0.078 & 0.086 & 0.093 & 0.097 & 0.101 \\
         & R-Ins & 0.033 & 0.048 & 0.058 & 0.072 & 0.079 & 0.088 & 0.094 & 0.098 & 0.102 & 0.108 \\
         & R-Sub & 0.033 & 0.048 & 0.062 & 0.072 & 0.075 & 0.081 & 0.085 & 0.093 & 0.100 & 0.108 \\
         & R-Sw & 0.032 & 0.048 & 0.054 & 0.067 & 0.074 & 0.081 & 0.088 & 0.093 & 0.102 & 0.107 \\
         & Rel & 0.016 & 0.021 & 0.023 & 0.029 & 0.030 & 0.035 & 0.040 & 0.043 & 0.050 & 0.052 \\
         & Split & 0.029 & 0.045 & 0.053 & 0.065 & 0.071 & 0.078 & 0.083 & 0.091 & 0.096 & 0.099 \\
        \midrule
        \multirow[t]{11}{*}{HerBERT} & Diac & 0.003 & 0.005 & 0.005 & 0.006 & 0.006 & 0.007 & 0.009 & 0.010 & 0.012 & 0.014 \\
         & Key & 0.026 & 0.041 & 0.049 & 0.055 & 0.060 & 0.062 & 0.067 & 0.077 & 0.081 & 0.087 \\
         & OCR & 0.027 & 0.038 & 0.053 & 0.057 & 0.063 & 0.067 & 0.073 & 0.078 & 0.086 & 0.095 \\
         & Ort & 0.006 & 0.008 & 0.010 & 0.011 & 0.013 & 0.015 & 0.016 & 0.018 & 0.021 & 0.026 \\
         & R-Del & 0.019 & 0.032 & 0.036 & 0.042 & 0.043 & 0.050 & 0.053 & 0.058 & 0.062 & 0.070 \\
         & R-Ins & 0.022 & 0.037 & 0.043 & 0.052 & 0.058 & 0.063 & 0.067 & 0.070 & 0.075 & 0.080 \\
         & R-Sub & 0.025 & 0.039 & 0.048 & 0.057 & 0.065 & 0.069 & 0.076 & 0.081 & 0.086 & 0.095 \\
         & R-Sw & 0.020 & 0.033 & 0.041 & 0.045 & 0.051 & 0.053 & 0.057 & 0.060 & 0.064 & 0.071 \\
         & Rel & 0.010 & 0.016 & 0.018 & 0.022 & 0.024 & 0.027 & 0.032 & 0.035 & 0.039 & 0.043 \\
         & Split & 0.018 & 0.027 & 0.033 & 0.037 & 0.040 & 0.044 & 0.047 & 0.051 & 0.055 & 0.060 \\
        \midrule
        \multirow[t]{11}{*}{RoBERTa} & Diac & 0.006 & 0.007 & 0.008 & 0.010 & 0.011 & 0.015 & 0.021 & 0.025 & 0.030 & 0.037 \\
         & Key & 0.031 & 0.049 & 0.060 & 0.071 & 0.080 & 0.092 & 0.100 & 0.112 & 0.123 & 0.132 \\
         & OCR & 0.037 & 0.058 & 0.073 & 0.085 & 0.092 & 0.107 & 0.115 & 0.124 & 0.135 & 0.142 \\
         & Ort & 0.007 & 0.010 & 0.012 & 0.014 & 0.018 & 0.025 & 0.030 & 0.037 & 0.044 & 0.051 \\
         & R-Del & 0.027 & 0.043 & 0.055 & 0.061 & 0.068 & 0.083 & 0.090 & 0.100 & 0.108 & 0.118 \\
         & R-Ins & 0.029 & 0.047 & 0.058 & 0.068 & 0.077 & 0.091 & 0.100 & 0.110 & 0.120 & 0.128 \\
         & R-Sub & 0.031 & 0.046 & 0.059 & 0.068 & 0.076 & 0.088 & 0.099 & 0.111 & 0.120 & 0.130 \\
         & R-Sw & 0.030 & 0.043 & 0.059 & 0.071 & 0.077 & 0.090 & 0.103 & 0.109 & 0.119 & 0.130 \\
         & Rel & 0.011 & 0.017 & 0.021 & 0.024 & 0.027 & 0.033 & 0.038 & 0.044 & 0.051 & 0.059 \\
         & Split & 0.027 & 0.042 & 0.053 & 0.063 & 0.071 & 0.080 & 0.083 & 0.091 & 0.098 & 0.109 \\
        \midrule
        \bottomrule
    \end{tabular}
    }
    \label{tab:smoothgrad_asr_per_model}
\end{table*}

\begin{table*}[!ht]
    \centering
    \caption{ASR for different numbers of words changed when using IntegratedGradients as attribution method, aggregated over datasets.}
    \resizebox{\textwidth}{!}{
    \begin{tabular}{ll|rrrrrrrrrr}
    \toprule
        \textbf{Model} & \textbf{Aug} & 1 & 2 & 3 & 4 & 5 & 6 & 7 & 8 & 9 & 10  \\
        \midrule
        \multirow[t]{11}{*}{PolBERT} & Diac & 0.003 & 0.006 & 0.006 & 0.008 & 0.010 & 0.011 & 0.013 & 0.015 & 0.019 & 0.021 \\
         & Key & 0.044 & 0.073 & 0.088 & 0.101 & 0.113 & 0.126 & 0.132 & 0.138 & 0.143 & 0.149 \\
         & OCR & 0.045 & 0.080 & 0.098 & 0.118 & 0.134 & 0.147 & 0.160 & 0.165 & 0.170 & 0.174 \\
         & Ort & 0.009 & 0.014 & 0.016 & 0.019 & 0.023 & 0.025 & 0.030 & 0.034 & 0.037 & 0.041 \\
         & R-Del & 0.043 & 0.068 & 0.085 & 0.104 & 0.119 & 0.130 & 0.134 & 0.142 & 0.149 & 0.155 \\
         & R-Ins & 0.045 & 0.073 & 0.085 & 0.103 & 0.117 & 0.124 & 0.131 & 0.145 & 0.150 & 0.155 \\
         & R-Sub & 0.043 & 0.072 & 0.086 & 0.105 & 0.117 & 0.129 & 0.137 & 0.144 & 0.145 & 0.148 \\
         & R-Sw & 0.042 & 0.068 & 0.082 & 0.099 & 0.110 & 0.124 & 0.138 & 0.142 & 0.149 & 0.157 \\
         & Rel & 0.015 & 0.027 & 0.031 & 0.040 & 0.044 & 0.050 & 0.055 & 0.057 & 0.064 & 0.070 \\
         & Split & 0.048 & 0.074 & 0.087 & 0.104 & 0.120 & 0.129 & 0.139 & 0.146 & 0.150 & 0.153 \\
        \midrule
        \multirow[t]{11}{*}{HerBERT} & Diac & 0.003 & 0.003 & 0.004 & 0.005 & 0.005 & 0.007 & 0.009 & 0.011 & 0.012 & 0.014 \\
         & Key & 0.029 & 0.046 & 0.061 & 0.072 & 0.082 & 0.090 & 0.098 & 0.109 & 0.118 & 0.126 \\
         & OCR & 0.028 & 0.048 & 0.058 & 0.067 & 0.078 & 0.085 & 0.091 & 0.101 & 0.111 & 0.115 \\
         & Ort & 0.005 & 0.009 & 0.013 & 0.014 & 0.018 & 0.019 & 0.022 & 0.024 & 0.026 & 0.029 \\
         & R-Del & 0.023 & 0.036 & 0.046 & 0.055 & 0.060 & 0.070 & 0.078 & 0.082 & 0.088 & 0.095 \\
         & R-Ins & 0.026 & 0.041 & 0.054 & 0.065 & 0.073 & 0.078 & 0.087 & 0.093 & 0.097 & 0.102 \\
         & R-Sub & 0.026 & 0.043 & 0.057 & 0.070 & 0.081 & 0.091 & 0.100 & 0.112 & 0.117 & 0.125 \\
         & R-Sw & 0.024 & 0.040 & 0.049 & 0.059 & 0.065 & 0.075 & 0.081 & 0.083 & 0.089 & 0.099 \\
         & Rel & 0.010 & 0.018 & 0.026 & 0.032 & 0.034 & 0.037 & 0.042 & 0.043 & 0.044 & 0.048 \\
         & Split & 0.019 & 0.029 & 0.039 & 0.046 & 0.050 & 0.055 & 0.062 & 0.073 & 0.077 & 0.083 \\
        \midrule
        \multirow[t]{11}{*}{RoBERTa} & Diac & 0.003 & 0.006 & 0.008 & 0.010 & 0.013 & 0.017 & 0.021 & 0.025 & 0.031 & 0.037 \\
         & Key & 0.045 & 0.079 & 0.101 & 0.124 & 0.143 & 0.160 & 0.175 & 0.195 & 0.208 & 0.219 \\
         & OCR & 0.055 & 0.093 & 0.119 & 0.136 & 0.154 & 0.171 & 0.188 & 0.207 & 0.219 & 0.230 \\
         & Ort & 0.009 & 0.016 & 0.019 & 0.023 & 0.029 & 0.035 & 0.042 & 0.048 & 0.055 & 0.062 \\
         & R-Del & 0.040 & 0.070 & 0.094 & 0.115 & 0.131 & 0.148 & 0.164 & 0.177 & 0.192 & 0.209 \\
         & R-Ins & 0.043 & 0.076 & 0.099 & 0.120 & 0.137 & 0.153 & 0.170 & 0.187 & 0.204 & 0.212 \\
         & R-Sub & 0.047 & 0.083 & 0.102 & 0.127 & 0.143 & 0.160 & 0.175 & 0.194 & 0.209 & 0.219 \\
         & R-Sw & 0.040 & 0.074 & 0.094 & 0.116 & 0.138 & 0.158 & 0.174 & 0.194 & 0.208 & 0.218 \\
         & Rel & 0.016 & 0.026 & 0.033 & 0.041 & 0.049 & 0.059 & 0.066 & 0.073 & 0.083 & 0.089 \\
         & Split & 0.037 & 0.069 & 0.089 & 0.105 & 0.117 & 0.134 & 0.148 & 0.165 & 0.178 & 0.184 \\
        \midrule
        \bottomrule
    \end{tabular}
    }
    \label{tab:ig_asr_per_model}
\end{table*}

\begin{table*}[!ht]
    \centering
    \caption{ASR for different numbers of words changed when using Attention Rollout as attribution method, aggregated over datasets.}
    \resizebox{\textwidth}{!}{
    \begin{tabular}{ll|rrrrrrrrrr}
    \toprule
        \textbf{Model} & \textbf{Aug} & 1 & 2 & 3 & 4 & 5 & 6 & 7 & 8 & 9 & 10  \\
        \midrule
        \multirow[t]{11}{*}{PolBERT} & Diac & 0.004 & 0.005 & 0.006 & 0.007 & 0.009 & 0.011 & 0.013 & 0.015 & 0.019 & 0.021 \\
         & Key & 0.030 & 0.057 & 0.070 & 0.090 & 0.111 & 0.127 & 0.143 & 0.156 & 0.163 & 0.171 \\
         & OCR & 0.032 & 0.063 & 0.082 & 0.100 & 0.117 & 0.138 & 0.153 & 0.167 & 0.181 & 0.190 \\
         & Ort & 0.004 & 0.008 & 0.012 & 0.014 & 0.018 & 0.022 & 0.025 & 0.030 & 0.034 & 0.038 \\
         & R-Del & 0.025 & 0.048 & 0.064 & 0.084 & 0.106 & 0.122 & 0.143 & 0.151 & 0.165 & 0.171 \\
         & R-Ins & 0.033 & 0.058 & 0.074 & 0.092 & 0.110 & 0.127 & 0.138 & 0.147 & 0.159 & 0.167 \\
         & R-Sub & 0.032 & 0.055 & 0.071 & 0.089 & 0.109 & 0.127 & 0.139 & 0.152 & 0.164 & 0.173 \\
         & R-Sw & 0.029 & 0.052 & 0.066 & 0.091 & 0.107 & 0.130 & 0.144 & 0.150 & 0.163 & 0.177 \\
         & Rel & 0.004 & 0.008 & 0.014 & 0.023 & 0.030 & 0.039 & 0.048 & 0.053 & 0.058 & 0.062 \\
         & Split & 0.026 & 0.051 & 0.064 & 0.083 & 0.106 & 0.124 & 0.139 & 0.149 & 0.162 & 0.171 \\
        \midrule
        \multirow[t]{11}{*}{HerBERT} & Diac & 0.003 & 0.003 & 0.004 & 0.005 & 0.006 & 0.006 & 0.008 & 0.009 & 0.010 & 0.013 \\
         & Key & 0.024 & 0.049 & 0.068 & 0.077 & 0.086 & 0.099 & 0.105 & 0.115 & 0.123 & 0.130 \\
         & OCR & 0.026 & 0.051 & 0.065 & 0.078 & 0.094 & 0.100 & 0.109 & 0.114 & 0.117 & 0.120 \\
         & Ort & 0.005 & 0.006 & 0.009 & 0.009 & 0.011 & 0.013 & 0.015 & 0.018 & 0.019 & 0.022 \\
         & R-Del & 0.018 & 0.033 & 0.041 & 0.050 & 0.053 & 0.060 & 0.066 & 0.072 & 0.076 & 0.087 \\
         & R-Ins & 0.024 & 0.044 & 0.056 & 0.070 & 0.075 & 0.082 & 0.090 & 0.098 & 0.102 & 0.107 \\
         & R-Sub & 0.028 & 0.049 & 0.066 & 0.078 & 0.085 & 0.099 & 0.102 & 0.109 & 0.119 & 0.125 \\
         & R-Sw & 0.014 & 0.033 & 0.042 & 0.049 & 0.057 & 0.064 & 0.070 & 0.075 & 0.081 & 0.089 \\
         & Rel & 0.008 & 0.011 & 0.015 & 0.019 & 0.024 & 0.026 & 0.031 & 0.035 & 0.038 & 0.045 \\
         & Split & 0.017 & 0.033 & 0.041 & 0.048 & 0.059 & 0.064 & 0.072 & 0.072 & 0.077 & 0.079 \\
        \midrule
        \multirow[t]{11}{*}{RoBERTa} & Diac & 0.003 & 0.005 & 0.006 & 0.007 & 0.010 & 0.015 & 0.020 & 0.025 & 0.030 & 0.036 \\
         & Key & 0.027 & 0.048 & 0.060 & 0.072 & 0.090 & 0.113 & 0.132 & 0.140 & 0.158 & 0.174 \\
         & OCR & 0.036 & 0.065 & 0.082 & 0.099 & 0.113 & 0.125 & 0.142 & 0.154 & 0.171 & 0.180 \\
         & Ort & 0.008 & 0.012 & 0.013 & 0.016 & 0.023 & 0.030 & 0.037 & 0.044 & 0.052 & 0.057 \\
         & R-Del & 0.022 & 0.036 & 0.046 & 0.063 & 0.081 & 0.103 & 0.119 & 0.130 & 0.147 & 0.158 \\
         & R-Ins & 0.028 & 0.047 & 0.059 & 0.075 & 0.090 & 0.113 & 0.129 & 0.142 & 0.157 & 0.168 \\
         & R-Sub & 0.030 & 0.048 & 0.063 & 0.075 & 0.094 & 0.110 & 0.128 & 0.142 & 0.157 & 0.167 \\
         & R-Sw & 0.021 & 0.039 & 0.056 & 0.065 & 0.078 & 0.100 & 0.121 & 0.137 & 0.155 & 0.170 \\
         & Rel & 0.004 & 0.008 & 0.013 & 0.015 & 0.024 & 0.033 & 0.042 & 0.052 & 0.060 & 0.071 \\
         & Split & 0.024 & 0.040 & 0.052 & 0.065 & 0.080 & 0.102 & 0.119 & 0.134 & 0.150 & 0.156 \\
        \midrule
        \bottomrule
    \end{tabular}
    }
    \label{tab:attn_asr_per_model}
\end{table*}

\begin{table*}[!ht]
    \centering
    \caption{ASR for different numbers of words changed when using Attention Gradient Rollout as attribution method, aggregated over datasets.}
    \resizebox{\textwidth}{!}{
    \begin{tabular}{ll|rrrrrrrrrr}
    \toprule
        \textbf{Model} & \textbf{Aug} & 1 & 2 & 3 & 4 & 5 & 6 & 7 & 8 & 9 & 10  \\
        \midrule
        \multirow[t]{11}{*}{PolBERT} & Diac & 0.006 & 0.007 & 0.008 & 0.012 & 0.014 & 0.015 & 0.018 & 0.019 & 0.022 & 0.026 \\
         & Key & 0.070 & 0.106 & 0.132 & 0.153 & 0.176 & 0.193 & 0.200 & 0.208 & 0.220 & 0.230 \\
         & OCR & 0.076 & 0.122 & 0.145 & 0.169 & 0.188 & 0.201 & 0.215 & 0.222 & 0.231 & 0.239 \\
         & Ort & 0.010 & 0.014 & 0.020 & 0.024 & 0.027 & 0.029 & 0.033 & 0.036 & 0.040 & 0.045 \\
         & R-Del & 0.062 & 0.104 & 0.131 & 0.157 & 0.168 & 0.181 & 0.194 & 0.204 & 0.215 & 0.224 \\
         & R-Ins & 0.066 & 0.106 & 0.129 & 0.147 & 0.165 & 0.179 & 0.188 & 0.200 & 0.212 & 0.219 \\
         & R-Sub & 0.072 & 0.115 & 0.137 & 0.158 & 0.180 & 0.188 & 0.206 & 0.216 & 0.221 & 0.228 \\
         & R-Sw & 0.061 & 0.099 & 0.131 & 0.151 & 0.168 & 0.186 & 0.197 & 0.204 & 0.216 & 0.226 \\
         & Rel & 0.019 & 0.039 & 0.049 & 0.060 & 0.065 & 0.071 & 0.076 & 0.078 & 0.084 & 0.089 \\
         & Split & 0.064 & 0.097 & 0.127 & 0.155 & 0.172 & 0.185 & 0.196 & 0.211 & 0.218 & 0.228 \\
        \midrule
        \multirow[t]{11}{*}{HerBERT} & Diac & 0.003 & 0.004 & 0.005 & 0.005 & 0.006 & 0.008 & 0.009 & 0.011 & 0.014 & 0.016 \\
         & Key & 0.033 & 0.056 & 0.073 & 0.083 & 0.096 & 0.107 & 0.114 & 0.122 & 0.130 & 0.139 \\
         & OCR & 0.034 & 0.053 & 0.075 & 0.085 & 0.099 & 0.110 & 0.115 & 0.127 & 0.132 & 0.141 \\
         & Ort & 0.005 & 0.008 & 0.009 & 0.011 & 0.013 & 0.014 & 0.018 & 0.020 & 0.022 & 0.024 \\
         & R-Del & 0.026 & 0.041 & 0.056 & 0.066 & 0.076 & 0.078 & 0.085 & 0.093 & 0.100 & 0.104 \\
         & R-Ins & 0.026 & 0.044 & 0.059 & 0.067 & 0.074 & 0.083 & 0.091 & 0.100 & 0.104 & 0.111 \\
         & R-Sub & 0.031 & 0.054 & 0.070 & 0.082 & 0.088 & 0.102 & 0.115 & 0.124 & 0.130 & 0.144 \\
         & R-Sw & 0.027 & 0.042 & 0.055 & 0.062 & 0.071 & 0.082 & 0.093 & 0.099 & 0.105 & 0.115 \\
         & Rel & 0.007 & 0.015 & 0.020 & 0.024 & 0.029 & 0.033 & 0.038 & 0.041 & 0.044 & 0.048 \\
         & Split & 0.023 & 0.040 & 0.054 & 0.062 & 0.069 & 0.076 & 0.082 & 0.086 & 0.091 & 0.096 \\
        \midrule
        \multirow[t]{11}{*}{RoBERTa} & Diac & 0.005 & 0.008 & 0.010 & 0.011 & 0.014 & 0.019 & 0.024 & 0.028 & 0.033 & 0.039 \\
         & Key & 0.038 & 0.067 & 0.098 & 0.119 & 0.138 & 0.156 & 0.169 & 0.184 & 0.204 & 0.216 \\
         & OCR & 0.045 & 0.087 & 0.115 & 0.132 & 0.152 & 0.169 & 0.184 & 0.198 & 0.218 & 0.231 \\
         & Ort & 0.008 & 0.012 & 0.020 & 0.022 & 0.026 & 0.033 & 0.043 & 0.049 & 0.057 & 0.063 \\
         & R-Del & 0.035 & 0.060 & 0.090 & 0.112 & 0.129 & 0.154 & 0.172 & 0.183 & 0.198 & 0.211 \\
         & R-Ins & 0.034 & 0.061 & 0.089 & 0.105 & 0.127 & 0.150 & 0.166 & 0.178 & 0.194 & 0.205 \\
         & R-Sub & 0.037 & 0.070 & 0.099 & 0.121 & 0.139 & 0.162 & 0.174 & 0.185 & 0.203 & 0.215 \\
         & R-Sw & 0.035 & 0.066 & 0.096 & 0.119 & 0.134 & 0.154 & 0.171 & 0.187 & 0.208 & 0.223 \\
         & Rel & 0.010 & 0.019 & 0.029 & 0.037 & 0.044 & 0.054 & 0.062 & 0.072 & 0.082 & 0.092 \\
         & Split & 0.032 & 0.059 & 0.085 & 0.100 & 0.116 & 0.133 & 0.149 & 0.161 & 0.178 & 0.190 \\
        \midrule
        \bottomrule
    \end{tabular}
    }    
    \label{tab:attn_grad_asr_per_model}
\end{table*}

\begin{figure*}[!ht]
    \centering
        \centering
        \includegraphics[width=\linewidth]{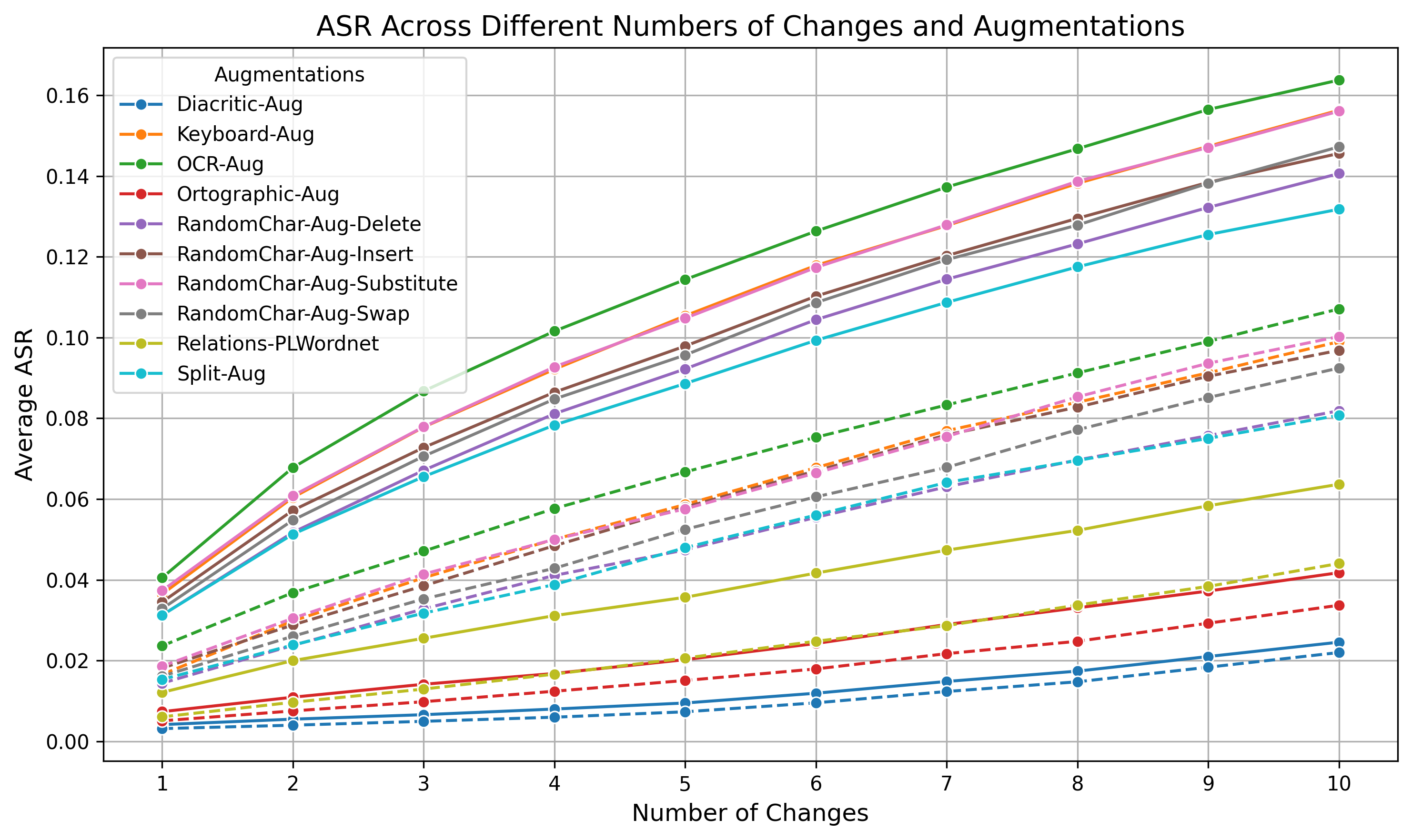}
    \caption{Relation of ASR on smaller LMs and number of perturbed words for different attribution methods. The dotted lines show ASR when selecting random words for perturbation instead of words based on word importance.}
    \label{fig:ASR_for_multiple_changes_LMs}
\end{figure*}

Figure~\ref{fig:ASR_for_multiple_changes_LMs} indicates that using attribution methods (targeted attack) significantly improves the effectiveness of the attack on language models compared to selecting random words during the attack.

\begin{figure*}[!ht]
    \centering
        \centering
        \includegraphics[width=\linewidth]{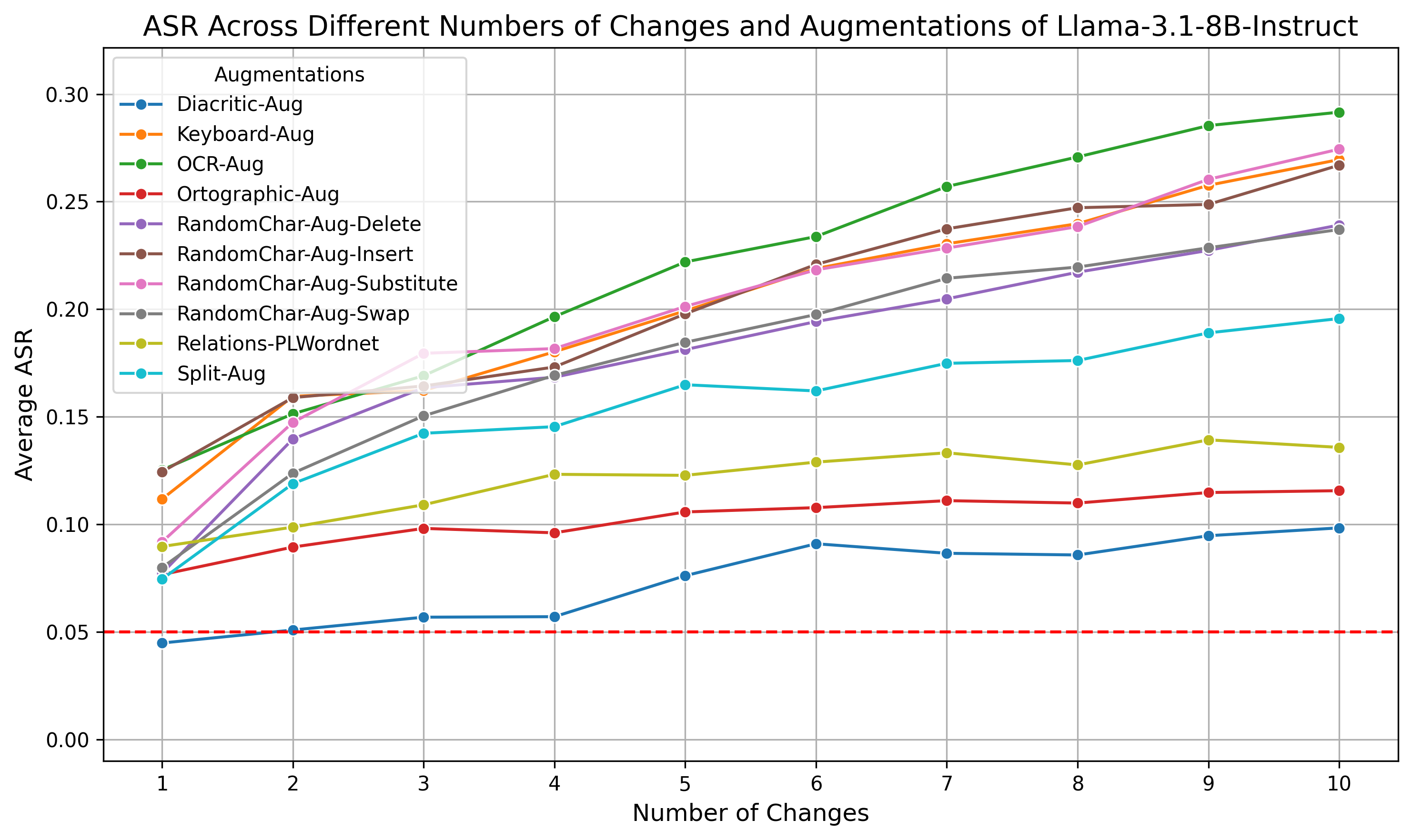}
    \caption{Relation of ASR on Llama and the number of perturbed words for different attribution methods. The dotted line indicates the robustness cutoff above which the model is considered non-robust to simple perturbations.}
    \label{fig:ASR_for_multiple_changes_llama}
\end{figure*}

\begin{figure*}[!ht]
    \centering
        \centering
        \includegraphics[width=\linewidth]{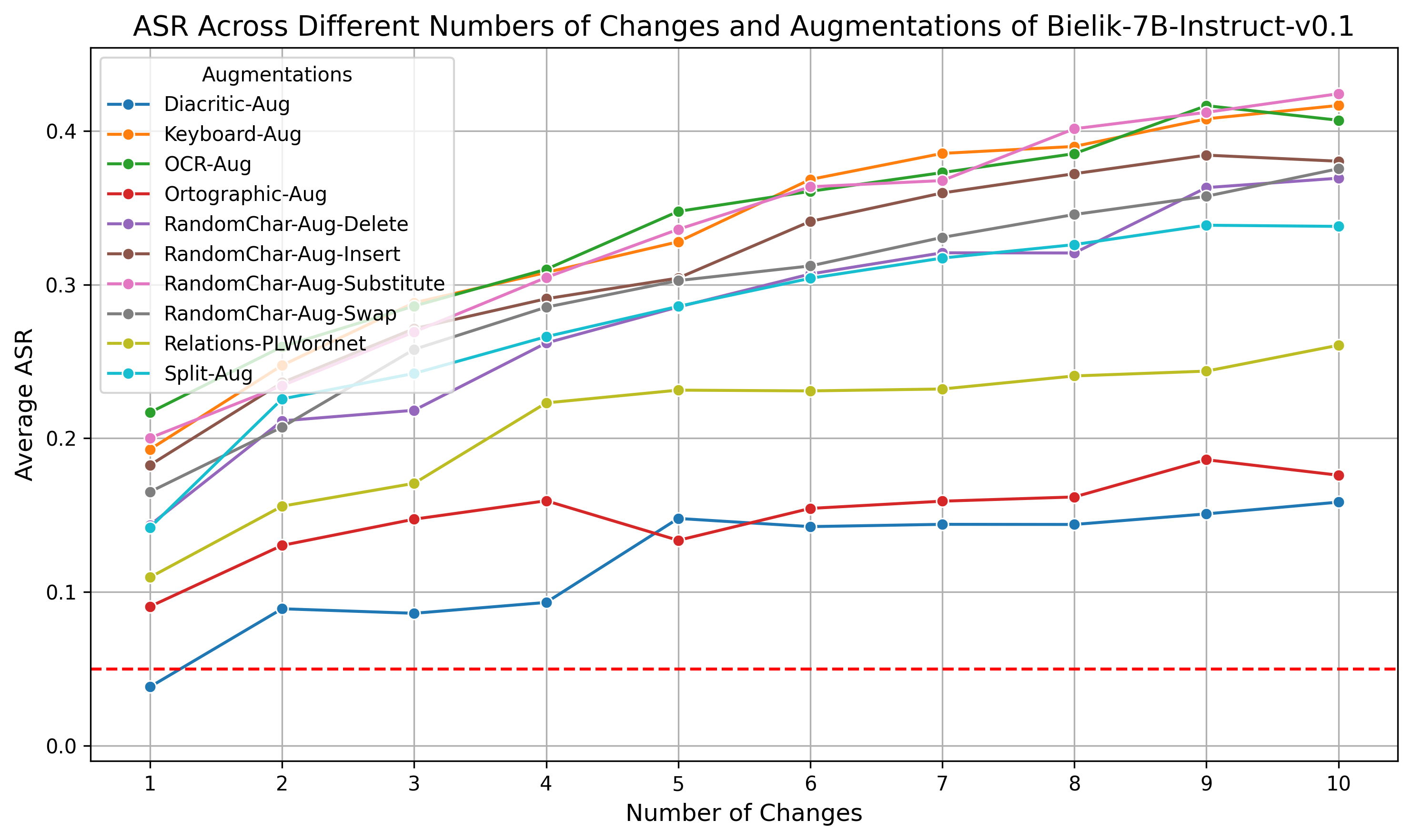}
    \caption{Relation of ASR on Bielik and the number of perturbed words for different attribution methods. The dotted line indicates the robustness cutoff above which the model is considered non-robust to simple perturbations.}
    \label{fig:ASR_for_multiple_changes_bielik}
\end{figure*}

Figures~\ref{fig:ASR_for_multiple_changes_llama},\ref{fig:ASR_for_multiple_changes_bielik} show robustness checks applied to Llama nad Bielik.

% Bibliography

\end{document}